\documentclass[11pt]{article}

\usepackage[final]{acl}
\usepackage{amsmath}
\usepackage{amssymb}
\usepackage{mathtools}
\usepackage{amsthm}
\usepackage{tabularx}  % 导言区加入
\usepackage{booktabs}  % 保持美观表格线
\usepackage{array}
\usepackage[table]{xcolor}
\usepackage{multirow}
\usepackage{multicol} 
\usepackage{array}    
\usepackage{algorithm}
\usepackage{algpseudocode}
\usepackage{xcolor}
\usepackage{setspace}
\algrenewcommand\algorithmicrequire{\textbf{Input:}}
\algrenewcommand\algorithmicensure{\textbf{Output:}}
\usepackage{tabularx}
\usepackage{pifont}
\usepackage[utf8]{inputenc}
\usepackage{caption}  
\usepackage{subcaption} 
\usepackage{adjustbox} 
\usepackage{afterpage}
\usepackage[most]{tcolorbox}
\usepackage{times}
\usepackage{latexsym}

\usepackage[T1]{fontenc}
\usepackage[utf8]{inputenc}

\usepackage{microtype}

\usepackage{inconsolata}

\usepackage{graphicx}

\title{PreResQ-R1: Response-Preference Disentangled Ranking-and-Scoring Reinforcement Optimization for Robust Visual Quality Assessment}
\author{
  \textbf{Zehui Feng\textsuperscript{1}},
  \textbf{Weichuan Wang\textsuperscript{1}},
  \textbf{Xiaohan Chen\textsuperscript{1}},
  \textbf{Ting Han\textsuperscript{1,2,3,*}} \\
  \textsuperscript{1}School of Design, Shanghai Jiao Tong University, Shanghai, China \\
  \textsuperscript{2}Institute of Medical Robotics, Shanghai Jiao Tong University, Shanghai, China \\
  \textsuperscript{3}Innovation Center of Yangtze River Delta, Zhejiang University, Jiaxing, Zhejiang \\
  \texttt{\{fzh\_sjtu, weichuanwang, chen\_11, hanting\}@sjtu.edu.cn}
}

\begin{document}
\maketitle

\begin{abstract}
Visual Quality Assessment (QA) in multimodal large language models (MLLMs) jointly predicts absolute quality scores and relative perceptual rankings. Existing methods optimize score regression or ranking within a unified formulation, leading to unstable predictions and weak generalization under stochastic generation. We find this instability is amplified by response variability in MLLM reasoning, where stochastic decoding produces inconsistent score distributions and degrades cross-dataset robustness. This indicates QA depends not only on preference alignment, but also on response-level stability. We propose PreResQ-R1, a Response-Preference Disentangled Ranking-and-Scoring Reinforcement Optimization framework. It decouples response stabilization and preference learning, with a dual-branch reward modeling intra-sample consistency and inter-sample score-ranking alignment. For video QA, we introduce temporal-spatial evidence aggregation without dense motion modeling. PreResQ-R1 achieves state-of-the-art performance, outperforming comparable methods by 5.60\% on IQA benchmarks and 2.53\% on VQA benchmarks.
\end{abstract}
\section{Introduction}
\begin{figure*}[t] 
\centering 
\vspace{-3mm} 
\includegraphics[width=1\linewidth]{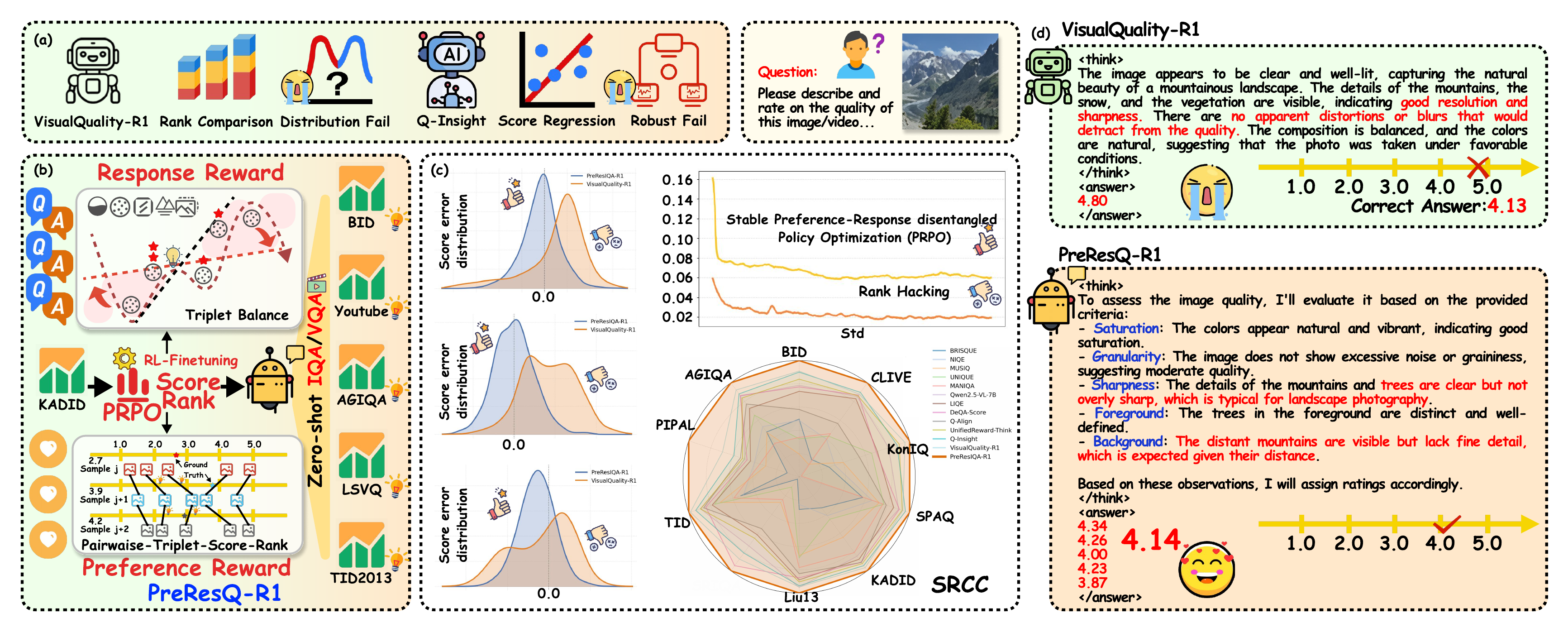} 
\vspace{-8mm} 
\caption{\textbf{Overview Performance}. (a) Existing score/ranking framework assign minimal difference, which results in distribution fall or robustness fail. (b) PreResQ-R1 focus on fine-grained response-ranking disentangled RL. (c) PreResQ-R1 enables state-of-the-art performance and stable image quality assessment. (d) typical comparisons with VisualQuality-R1 show that PreResQ-R1 achieves more accurate image quality description and scoring. } 
\label{fig:overall} 
\vspace{-5mm} 
\end{figure*}
Visual Quality Assessment (QA) aims to predict human perceptual judgments of visual fidelity without reference images~\cite{1284395}, and is fundamental to applications such as image enhancement~\cite{chen2024promptbasedtesttimerealimage}, image generation~\cite{10771738}, and computational photography~\cite{cai2024phocolensphotorealisticconsistentreconstruction}. The absence of reference images makes perceptual inference challenging in robustness and interpretability. A key limitation lies in the joint optimization of absolute score regression and relative preference ranking within a single objective \cite{wu2025visualqualityr1reasoninginducedimagequality}. Regression-based methods provide calibrated scores but are sensitive to dataset-specific distributions, leading to poor cross-dataset generalization~\cite{9710973, you2025teachinglargelanguagemodels}, while ranking-based methods improve relative ordering~\cite{ying2019patchespicturespaq2piqmapping, liu2017rankiqalearningrankingsnoreference} but lack absolute supervision, resulting in suboptimal calibration.

Recent multimodal large language model (MLLM) based QA methods attempt to address this limitation by incorporating reasoning through chain-of-thought supervision and reinforcement learning (RL) \cite{touvron2023llamaopenefficientfoundation, bai2025qwen25vltechnicalreport, wang2025unifiedmultimodalchainofthoughtreward}. However, they still rely on unified scalar rewards without explicitly disentangling perceptual factors, and therefore fail to resolve the underlying objective mismatch. As a result, RL optimization suffers from noisy and unstable gradients across stochastic generations, leading to slow convergence and brittle generalization, even when advanced policy optimization schemes are employed~\cite{deepseekai2025deepseekr1incentivizingreasoningcapability}.

Importantly, this instability is not merely an optimization artifact \cite{li2026theory,li2026correcting,li2024deep}; it reflects an inherent property of QA under stochastic generation. Responses to the same input can vary substantially, directly altering the learned score distribution. Left unconstrained, such variability harms calibration and weakens cross-dataset generalization. By contrast, response-level rational consistency acts as an implicit regularizer, improving the stability and reliability of quality estimation. These findings suggest that beyond preference alignment, response stability is essential for distributional robustness under limited supervision.

Extending reasoning-oriented QA models to video further amplifies this misalignment between perceptual requirements and model design \cite{li2026ragtrack,li2026cadtrack}. Existing methods~\cite{cao2025vqathinkerexploringgeneralizableexplainable} often rely on explicit motion modeling or dense temporal sampling, incurring additional computational cost and limited alignment with MLLM-based IQA reasoning. In contrast, human perception of video quality aggregates sparse spatial and temporal cues rather than exhaustive motion reconstruction \cite{FENG2026104289, FENG2026103914}, highlighting the need for perceptually grounded evidence aggregation under practical input constraints \cite{zhang2024defending,ti2025towards,wu2025lanp,chen2026towards}.

Motivated by these observations, we propose \textbf{PreResQ-R1}, a \textbf{Preference-Response Disentangled Reinforcement optimization} framework for visual QA. PreResQ-R1 decomposes perceptual learning into response-level stability and preference-level alignment, and integrates them within a unified Group Relative Policy Optimization (GRPO) scheme with long-chain reasoning. In particular, the response dimension is explicitly modeled to regularize stochastic generation and stabilize the learned score distribution, which in turn improves robustness and cross-distribution generalization, while the preference dimension enforces perceptual ordering with magnitude-aware calibration. For video QA task, we adopt a temporal-spatial evidence aggregation strategy that aligns with reasoning constraints without architectural modifications. Our contributions are threefold:\\
1. We propose a disentangled RL formulation that explicitly models response-level stability and preference-level alignment via complementary ranking-aware rewards.\\
2. We design a fine-grained reward scheme that jointly optimize ranking consistency and scoring alignment through complementary pairwise and triplet ranking signals.\\
3. Extensive experiments demonstrate that PreResQ-R1 achieves consistent improvement across diverse IQA and VQA benchmarks, producing interpretable, human-aligned reasoning traces.
\section{Related Work}
\vspace{-2mm}
\subsection{Regression-based models}
Classical IQA and VQA methods predominantly formulate perceptual quality assessment as an absolute regression problem, mapping visual inputs to scalar scores based on signal fidelity assumptions or handcrafted natural scene statistics~\cite{20080522,1284395,6272356,6353522}.
The emergence of deep learning enables end-to-end quality regression with CNN- and Transformer-based architectures, yielding substantial performance gains across benchmarks~\cite{8576582,9710973,Wu_2023,wu2022fastvqaefficientendtoendvideo}.
Recent extensions to MLLMs incorporate pretrained multimodal representations to enhance semantic reasoning~\cite{hinglmmsvisual,li2025qinsightunderstandingimagequality}. However, these models entangle perceptual uncertainty with score prediction and lead to sensitivity to dataset-specific score distributions \cite{FENG2026114584, FENG2025103363}.

\vspace{-2mm}
\subsection{Ranking-based models}
\vspace{-1mm}
Ranking-based IQA and VQA methods formulate quality assessment as a relative ranking task, which better aligns with human perceptual judgment. Early methods including RankIQA~\cite{liu2017rankiqalearningrankingsnoreference} optimize ranking objectives via correlation-based losses. Recent MLLM-based approaches, such as C2Score~\cite{zhu2024adaptiveimagequalityassessment}, DeQA-Score~\cite{you2025teachinglargelanguagemodels}, LMM-PVQA~\cite{cao2025generalizedvideoqualityassessment}, and VisualQuality-R1~\cite{wu2025visualqualityr1reasoninginducedimagequality}, further leverage comparative supervision to enhance both perceptual consistency and reasoning interpretability. However, ranking-based methods inherently sacrifice absolute score calibration, making cross-dataset comparability challenging.

\vspace{-2mm}
\begin{figure*}[t]
    \includegraphics[width=1\linewidth]{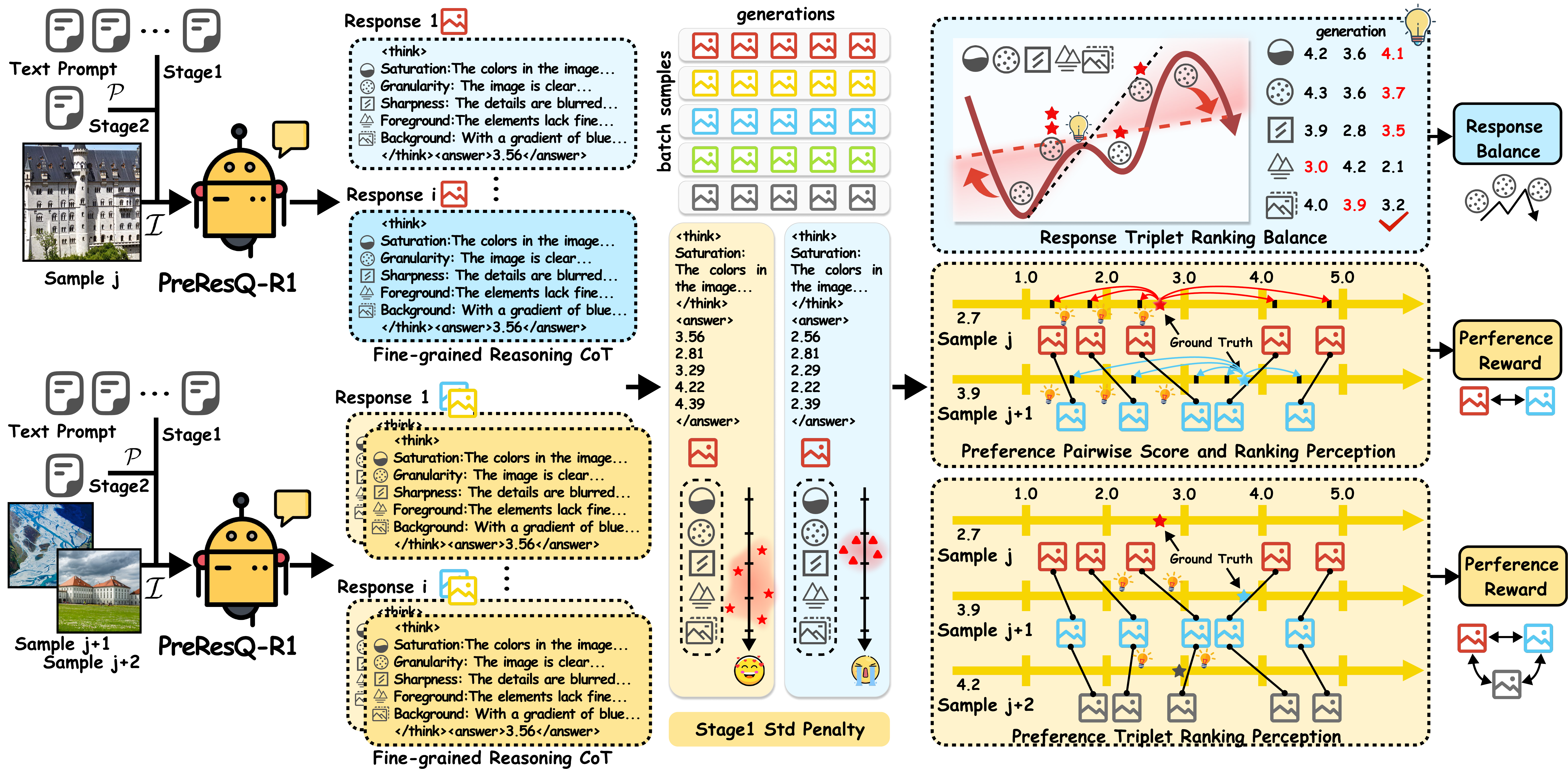}
    \vspace{-3mm}
    \caption{Overall training framework of PreResQ-R1 via reinforcement learning to rank and score. Given an sample batch ($\mathcal{I}$$_{j}$, $\mathcal{I}_{j+1}$,..., $\mathcal{I}_{j+B})$ with a shared text prompt $\mathcal{P}$, PreResQ-R1 generates $K$ responses. To quickly activate CoT differences and then access generation stability, we introduce the response penalty and fine-grained triplet-response balance reward. To jointly enhance the robustness of ranking and score ability, we introduce the preference pairwise-and-triplet score-and-ranking reward for GRPO.}
    \label{fig:model}
\vspace{-5mm}
\end{figure*}
\subsection{Reinforcement learning-based models}
\vspace{-1mm}
RL has recently emerged as a promising mechanism for aligning MLLMs with human perceptual preferences. Building on RLHF~\cite{ouyang2022traininglanguagemodelsfollow} and RLAIF~\cite{yu2024rlaifvopensourceaifeedback}, policy optimization frameworks such as GRPO~\cite{shao2024deepseekmathpushinglimitsmathematical} enable preference-driven refinement without explicit human annotations. This paradigm underlies models such as DeepSeek-R1-Zero~\cite{deepseekai2025deepseekr1incentivizingreasoningcapability}, R1-VL~\cite{zhang2025r1vllearningreasonmultimodal}, and Visual-RFT~\cite{liu2025visualrftvisualreinforcementfinetuning}.
In the context of QA, RL-based methods aim to bridge perceptual fidelity and reasoning. Q-Insight~\cite{li2025qinsightunderstandingimagequality} and VQ-Insight~\cite{zhang2025vqinsightteachingvlmsaigenerated} incorporate reward-based optimization for score prediction, while VisualQuality-R1~\cite{wu2025visualqualityr1reasoninginducedimagequality} poses IQA as a reasoning-guided ranking problem. Related works on video and preference alignment (e.g. VideoDPO~\cite{liu2024videodpoomnipreferencealignmentvideo}, VideoReward~\cite{liu2025improvingvideogenerationhuman}, and UnifiedReward~\cite{wang2025unifiedrewardmodelmultimodal}) demonstrate RL’s potential, while also exposing challenges in optimization stability and computational efficiency.
\section{Proposed Model}
\vspace{-2mm}
\subsection{Overview}
This work aims to incorporate long CoT reasoning into the reward MLLM's assessment process to enhance the robustness of reward signals. Evidence ~\cite{wu2025visualqualityr1reasoninginducedimagequality, hinglmmsvisual} shows that relying solely on score-based rewards often results in imbalanced score distributions, while rank-based rewards tend to be dataset-specific, leading to limitations in generalization across diverse tasks. The key challenge lies in devising an effective strategy to elicit and incentivize CoT capability. Therefore, we take the first step to proposed a preference-response disentangled multimodal CoT-based NR-QA model, PreResQ-R1 (Fig. \ref{fig:model}), to activates both score and rank convergence through fine-grained reasoning.
\subsection{Fine-grained CoT for IQA and VQA}
Given a text prompt $\mathcal{P}$ and an image or video $\mathcal{I}$ in an image/video, our goal is to fine-tune a pretrained MLLM with parameters $\theta$ and policy $\pi_{\theta}(\cdot|\mathcal{P},\mathcal{I})$ to generate quality reasoning trajectories $\tau$ that provide a comprehensive CoT $c$ along with an overall score $s \in [1, 5]$:
\begin{equation}
    \tau = (c, s) \sim \pi_{\theta}(\cdot | \mathcal{P}, \mathcal{I})
\end{equation}
We present the complete structured text prompt in Appendix \ref{text_prompt}. For each image-score pair $x_i = (\mathcal{P}_i, \mathcal{I}_i)$ in IQA, the MLLM generates detailed Chain-of-Thought (CoT) reasoning for five quality aspects: \textit{saturation}, \textit{granularity}, and \textit{sharpness}, which capture pixel-level properties, and \textit{foreground} and \textit{background}, which capture semantic-level information, followed by the corresponding numeric scores. For VQA, inspired by DeepSeek-OCR~\cite{wei2025deepseekocrcontextsopticalcompression}, we adopt a temporal-spatial representation compatible with IQA task. We uniformly sample $M$ frames to form a temporal-flow image summarizing long-range temporal evolution, and $N$ frames to construct local spatial-flow images capturing frame-level details. Although this formulation does not explicitly model fine-grained temporal continuity or motion-specific artifacts, it effectively exploits perceptual priors learned from IQA fine-tuning and enables scalable VQA under multilevel linear MLLM input constraints. Consequently, VQA is performed via structured aggregation of temporally distributed visual evidence rather than explicit motion reconstruction. For every training batch of image/video-score pairs $\{x_i, x_{i+1}, ..., x_{i+B}\}$, where $B$ is the minibatch size, we apply GRPO to generate $K$ quality predictions for $x_i$:
\begin{equation}
\small
    \tau(x_i) = [\tau_1(x_i), \tau_2(x_i), ..., \tau_K(x_i)]
\end{equation}
\subsection{Preference-Response disentangled Policy Optimization (PRPO)}
PRPO explicitly decomposes the policy optimization into two complementary components that jointly capture the essence of perceptual reasoning. The \textbf{Response Dimension} focuses on intra-sample consistency, evaluating how coherently the model reasons across multiple generations of the same sample. Meanwhile, the \textbf{Preference Dimension} models inter-sample relations, encouraging the predicted quality score to align with human MOS. In addition, a format reward $R_{for}$ is utilized to ensure specific answer format.
\subsubsection{Response-based ranking reward}
To accelerate early-stage convergence and avoid degenerate score collapse caused by highly implausible predictions, we introduce a response-based ranking reward that imposes a weak intra-sample consistency bias across stochastic generations. For each input sample $j$, given a triplet of stochastic predictions
$\{s_{j,d}^{l}, s_{j,d}^{m}, s_{j,d}^{n}\}$ on perceptual dimension $d$,
we use their median as a local reference and define the response-based ranking reward as:
\begin{equation}
\small
R_{RR}^{j,i}\!
=\!
\frac{1}{D}\sum_{d=1}^{D}\!
\exp\!\left(\!
-\left|
s_{j,d}^{i}\!
-\!
\mathrm{median}\!\left(\{s_{j,d}^{l}, s_{j,d}^{m}, s_{j,d}^{n}\}\right)
\right|
\right),
\end{equation}
This formulation suppresses extreme outliers while preserving meaningful score variability. To ensure numerical stability, triplets with a total absolute difference below 0.5 receive a reward of 1. RR serves as an early-phase optimization scaffold, guiding exploration away from implausible score regions, and naturally becomes secondary to preference-driven rewards as policy entropy decreases.

\subsubsection{Preference-based ranking-score reward}
\textbf{Pairwise Comparative Reward.}\\
Building upon the response-based ranking reward, which constrains stochastic generations of the same input into a stabilized response distribution, we further introduce a preference-based reward to enable effective inter-sample quality comparison. Under the RR-induced stabilization, multiple responses to the same input exhibit reduced variance and comparable relative ordering, providing a prerequisite for preference learning.
In quality assessment, purely score-based rewards are prone to distribution imbalance and score collapse, while rank-based rewards, although more robust to scale variations, often suffer from dataset-dependent ordinal bias and limited sensitivity to perceptual gaps. To balance these complementary strengths and weaknesses, we adopt a rank-score coupled preference formulation that jointly enforces ordinal correctness and magnitude awareness.
Specifically, for each sample $j$, we first perform an intra-sample ranking over its $K$ stochastic generations based on their predicted quality scores, yielding ordered statistics
$\hat{s}_j^{(1)} \le \hat{s}_j^{(2)} \le \cdots \le \hat{s}_j^{(K)}$.
This ranking serves as a normalization step that preserves relative quality structure learned under RR. Cross-sample preference learning is then conducted by comparing predictions at the same rank index $i$ across different samples. Rather than assuming absolute score comparability, this design leverages the RR-induced convergence property: predictions at the same order statistic index represent similar relative confidence levels across samples, enabling stable preference supervision.

Formally, let $\mathbf{G}_j$ denote the ground-truth MOS for sample $j$. For any sample pair $(l,m)$ and rank index $i$, we define an ordinal consistency indicator $\mathbb{I}$:
\begin{equation}
\small
C_{l,m}^{(i)}=
\mathbb{I}\!\left[
\mathrm{sign}\big(\hat{s}_l^{(i)}-\hat{s}_m^{(i)}\big)
=
\mathrm{sign}\big(\mathbf{G}_l-\mathbf{G}_m\big)
\right],
\end{equation}
which measures whether the $i$-th ranked predictions preserve the ground-truth ordering.
To complement ordinal supervision with perceptual sensitivity, we introduce a magnitude-aware alignment term:
\begin{equation}
\small
M_{l,m}^{(i)} =
\frac{|\mathbf{G}_l - \mathbf{G}_m|}
{|\hat{s}_l^{(i)} - \hat{s}_m^{(i)}|
+ |\hat{s}_l^{(i)} - \mathbf{G}_l|
+ |\hat{s}_m^{(i)} - \mathbf{G}_m|},
\end{equation}
The final preference-based ranking-score reward for generation $i$ of sample $l$ in minibatch size $B$ is defined as:
\begin{equation}
\small
R_{pair}^{l,i}
= \sum_{\substack{m=1\\m\neq l}}^B
\Bigg[\!
\sqrt{C_{l,m}^{(i)} e^{M_{l,m}^{(i)}}}
+\sqrt{(1-C_{l,m}^{(i)}) e^{-(1+M_{l,m}^{(i)})}}
\Bigg].
\end{equation}

\textbf{Triplet comparative reward:}
Pairwise ranking supervision enforces local ordinal correctness but does not guarantee global ranking consistency. In particular, pairwise-consistent predictions may still violate transitivity, leading to unstable or contradictory preference structures.
To complement pairwise constraints, we introduce a triplet-based comparative reward that explicitly enforces transitive consistency among three samples. By jointly evaluating all three pairwise relations within a triplet, the triplet reward regularizes the global ranking geometry and suppresses cyclic inconsistencies that cannot be resolved by pairwise supervision alone.
The pairwise and triplet rewards thus serve complementary roles: the former provides fine-grained local preference signals, while the latter acts as a structural constraint that stabilizes global ordering. We adopt a simple ordinal triplet formulation that avoids additional scale assumptions and preserves training stability.
For each anchor sample \(a\), we construct the set of all triplets \(\mathcal{T}_l = \{(l,m,n) \,|\, m,n \neq l\}\). For a single triplet \((l,m,n) \in \mathcal{T}_i\), the triplet reward is defined as:
\begin{equation}
\small
r_{tri}^{(l,m,n),i} =
\begin{cases}
1.0, & \text{if } C_{l,m}^{(i)} + C_{l,n}^{(i)} + C_{m,n}^{(i)} = 3,\\[1mm]
0.3, & \text{otherwise,}
\end{cases}
\end{equation}
where \(C_{l,m}^i \in \{0,1\}\) denotes the pairwise consistency between predictions \(s_l^k, s_m^k\) and ground-truth scores \(\mathbf{G}_l, \mathbf{G}_m\).  
The overall triplet reward for generation \(i\) and sample \(j\) is then computed as the mean over all triplets:
\begin{equation}
\small
R_{tri}^{j,i} = \frac{1}{|\mathcal{T}_j|} \sum_{(l,m,n) \in \mathcal{T}_j} r_{tri}^{(l,m,n),i}.
\end{equation}
\vspace{-6mm}
\subsection{Total objective}
\begin{table*}[t]
\centering
\caption{Performance comparison of PreResQ-R1 across KADID-10K IQA datasets.
For each dataset, SRCC (S) and PLCC (P) are reported.
\textcolor{red}{\textbf{Red}}, \textcolor{green}{\textbf{green}}, and \textcolor{blue}{\textbf{blue}} denote handcrafted, deep-learning-based, and MLLM-based methods.
\textbf{Bold} and \underline{underline} indicate the best and second best results. All reported results are obtained by averaging over ten independent runs with different random seeds.}
\vspace{-3mm}
\scriptsize
\renewcommand{\arraystretch}{1}
\setlength{\tabcolsep}{1.6pt}

\begin{adjustbox}{width=\textwidth}
\begin{tabular}{l cc cc cc cc cc cc cc cc cc cc cc}
\toprule
\multirow{2}{*}{Method}
& \multicolumn{2}{c}{BID}
& \multicolumn{2}{c}{CLIVE}
& \multicolumn{2}{c}{KonIQ}
& \multicolumn{2}{c}{SPAQ}
& \multicolumn{2}{c}{KADID}
& \multicolumn{2}{c}{Liu13}
& \multicolumn{2}{c}{SRIQA}
& \multicolumn{2}{c}{TID13}
& \multicolumn{2}{c}{PIPAL}
& \multicolumn{2}{c}{AGIQA}
& \multicolumn{2}{c}{Avg} \\
\cmidrule(lr){2-23}
& SRCC & PLCC & SRCC & PLCC & SRCC & PLCC & SRCC & PLCC & SRCC & PLCC & SRCC & PLCC & SRCC & PLCC & SRCC & PLCC & SRCC & PLCC & SRCC & PLCC & SRCC & PLCC \\
\midrule

\rowcolor{purple!10}
BRISQUE & 0.522 & 0.528 & 0.314 & 0.362 & 0.385 & 0.400 & 0.614 & 0.624 & 0.429 & 0.451 & 0.389 & 0.380 & 0.210 & 0.163 & 0.548 & 0.546 & 0.242 & 0.259 & 0.497 & 0.541 & 0.415 & 0.425 \\

\rowcolor{purple!10}
NIQE & 0.515 & 0.527 & 0.450 & 0.494 & 0.421 & 0.439 & 0.676 & 0.683 & 0.487 & 0.415 & 0.360 & 0.376 & 0.194 & 0.135 & 0.532 & 0.516 & 0.357 & 0.314 & 0.533 & 0.560 & 0.453 & 0.446 \\

\rowcolor{green!10}
MUSIQ & 0.327 & 0.280 & 0.284 & 0.325 & 0.473 & 0.435 & 0.720 & 0.666 & 0.647 & 0.622 & 0.656 & 0.563 & 0.309 & 0.283 & 0.670 & 0.695 & 0.317 & 0.347 & 0.494 & 0.434 & 0.490 & 0.456 \\

\rowcolor{green!10}
UNIQUE & 0.412 & 0.385 & 0.470 & 0.472 & 0.649 & 0.590 & 0.751 & 0.708 & 0.513 & 0.548 & 0.669 & 0.654 & 0.244 & 0.207 & 0.703 & 0.729 & 0.393 & 0.361 & 0.608 & 0.581 & 0.541 & 0.524 \\

\rowcolor{green!10}
MANIQA & 0.420 & 0.512 & 0.487 & 0.571 & 0.213 & 0.257 & 0.745 & 0.753 & 0.760 & 0.780 & 0.726 & 0.728 & 0.297 & 0.289 & 0.589 & 0.617 & 0.338 & 0.364 & 0.422 & 0.448 & 0.500 & 0.532 \\

\midrule
\rowcolor{blue!10}
Qwen2.5-VL & 0.711 & 0.725 & 0.733 & 0.760 & 0.754 & 0.810 & 0.848 & 0.854 & 0.787 & 0.806 & \underline{0.840} & 0.852 & 0.495 & 0.504 & 0.787 & 0.837 & 0.390 & 0.420 & 0.735 & 0.772 & 0.708 & 0.734 \\

\rowcolor{blue!10}
LIQE & 0.677 & 0.680 & 0.719 & 0.726 & 0.684 & 0.652 & 0.815 & 0.814 & 0.809 & 0.817 & 0.797 & 0.712 & 0.447 & 0.430 & 0.718 & 0.748 & 0.371 & 0.347 & 0.653 & 0.653 & 0.669 & 0.658 \\

\rowcolor{blue!10}
DeQA-Score & 0.702 & 0.743 & 0.743 & \underline{0.795} & 0.677 & 0.703 & 0.852 & 0.858 & 0.831 & 0.873 & 0.785 & 0.838 & \underline{0.579} & \underline{0.589} & 0.756 & 0.793 & 0.383 & 0.381 & 0.738 & 0.743 & 0.705 & 0.732 \\

\rowcolor{blue!10}
Q-Align & 0.576 & 0.651 & 0.554 & 0.643 & 0.573 & 0.612 & 0.767 & 0.779 & 0.832 & 0.862 & 0.761 & 0.802 & 0.421 & 0.459 & 0.769 & 0.794 & 0.406 & 0.381 & 0.682 & 0.694 & 0.634 & 0.668 \\

\rowcolor{blue!10}
UnifiedReward-T & 0.745 & 0.767 & 0.738 & 0.795 & 0.820 & 0.804 & 0.871 & 0.846 & 0.841 & 0.877 & 0.809 & 0.835 & 0.462 & 0.518 & 0.788 & \underline{0.873} & 0.399 & 0.421 & 0.722 & 0.745 & 0.720 & 0.748 \\

\rowcolor{blue!10}
Q-Insight & 0.784 & 0.796 & \underline{0.761} & \underline{0.795} & 0.806 & 0.829 & \underline{0.892} & 0.872 & 0.856 & \underline{0.881} & 0.831 & 0.857 & 0.525 & 0.542 & 0.816 & 0.851 & 0.429 & \textbf{0.485} & 0.749 & 0.794 & 0.745 & 0.770 \\

\rowcolor{blue!10}
VisualQuality-R1 & \underline{0.790} & \underline{0.806} & 0.750 & 0.794 & \underline{0.855} & \underline{0.840} & 0.875 & \underline{0.878} & \underline{0.871} & 0.821 & \underline{0.838} & \textbf{0.872} & 0.561 & 0.543 & \underline{0.848} & 0.871 & \underline{0.469} & 0.458 & \underline{0.805} & \textbf{0.843} & 0.768 & 0.773 \\

\rowcolor{blue!10}
PreResQ-R1 (Ours) & 
\textbf{0.849} & \textbf{0.839} &
\textbf{0.825} & \textbf{0.825} &
\textbf{0.884} & \textbf{0.887} &
\textbf{0.919} & \textbf{0.900} &
\textbf{0.894} & \textbf{0.892} &
\textbf{0.873} & \underline{0.864} &
\textbf{0.601} & \textbf{0.596} &
\textbf{0.858} & \textbf{0.877} &
\textbf{0.588} & \underline{0.431} &
\textbf{0.823} & \underline{0.809} &
\textbf{0.811} & \textbf{0.790} \\

\bottomrule
\end{tabular}
\end{adjustbox}

\vspace{-2mm}
\label{tab:comparison}
\end{table*}
We unify response-level fidelity and cross-sample preference supervision into a single scalar reward. For the $i$-th generation of sample $j$, the total reward is defined as
\begin{equation}
\small
R_{\mathrm{total}}^{j,i}
= R_{\mathrm{for}}^{j,i}
+ \alpha \, R_{\mathrm{RR}}^{j,i}
+ (1-\alpha)\big( \beta \, R_{\mathrm{pair}}^{j,i} + (1-\beta) \, R_{\mathrm{tri}}^{j,i} \big),
\end{equation}
where $\alpha,\! \beta\in [0,1]$.
This hierarchical design constrains individual responses to be perceptually faithful and distributionally stable while enforcing both local and global preference consistency across samples. The resulting objective aligns intra-response quality with inter-sample ordering, enabling stable reinforcement fine-tuning under stochastic generation.
The model is optimized by a regularized policy gradient objective. Detailed reinforcement learning formulations are provided in Appendix~\ref{app:rl}.
\subsection{Exploration-to-stability fine-tuning strategy}
To balance exploration and stable fine-tuning, we adopt a two-stage fine-tuning strategy. In the exploration stage ($\tau$), each sample is paired with $\mathbf{X}$ randomly sampled prompts, encouraging the model to generate diverse trajectories. For each generation, we extract the five-dimensional scores $\mathbf{s}_{j}^{(i)}$ and compute a dimension-wise standard deviation penalty:
\begin{equation}
\small
R_{\mathrm{std}}(\mathbf{s}_j^{(i)}) = 
\begin{cases}
\lambda_{\mathrm{std}} (\delta_{\min} - \sigma(\mathbf{s}_j^{(i)}),&\sigma(\mathbf{s}_j^{(i)}) < \delta_{\min},\\
0,&\text{otherwise,}
\end{cases}
\end{equation}
where $\mathbf{s}_j^{(i)} = [s_{j}^1, \dots, s_{j}^5]$ denotes the five-dimensional quality scores for the $i$-th generation of sample $j$, and $\sigma(\mathbf{s}_j^{(i)})$ is the standard deviation across these five scores. 
\section{Experiment}
\vspace{-2mm}
\subsection{Experimental setups}
For \textbf{IQA task}, PreResQ-R1 is trained exclusively on the synthetic KADID-10K and KonIQ dataset (6:2:2 split in content independence) and tested in a zero-shot setting across ten open-source general (real-world and AI-generated) image aesthetic assessment datasets: BID~\cite{5492198}, CLIVE~\cite{7327186}, KonIQ-10k~\cite{Hosu_2020}, SPAQ~\cite{9156490}, KADID~\cite{8743252}, Liu 13~\cite{10.1145/2508363.2508391}, SRIQA-Bench~\cite{chen2025generalizedimagequalityassessment}, AGIQA-3K~\cite{li2023agiqa3kopendatabaseaigenerated}, TID 2013~\cite{PONOMARENKO201557}, and PIPAL~\cite{gu2020pipallargescaleimagequality}.

For \textbf{VQA task}, PreResQ-R1 is trained exclusively on the LSVQ-28K \cite{Ying_2021} dataset and test in a zero-shot setting across five open-source in-domain UGC videos with in-the-wild distortions: LSVQ$_{\text{test}}$ \cite{Ying_2021}, KoNViD-1k \cite{Hosu_2020}, LIVE-VQC \cite{8463581}, YouTube-UGC\cite{Wang_2019}.

\begin{figure*}[t]
    \centering
    \includegraphics[width=1\linewidth]{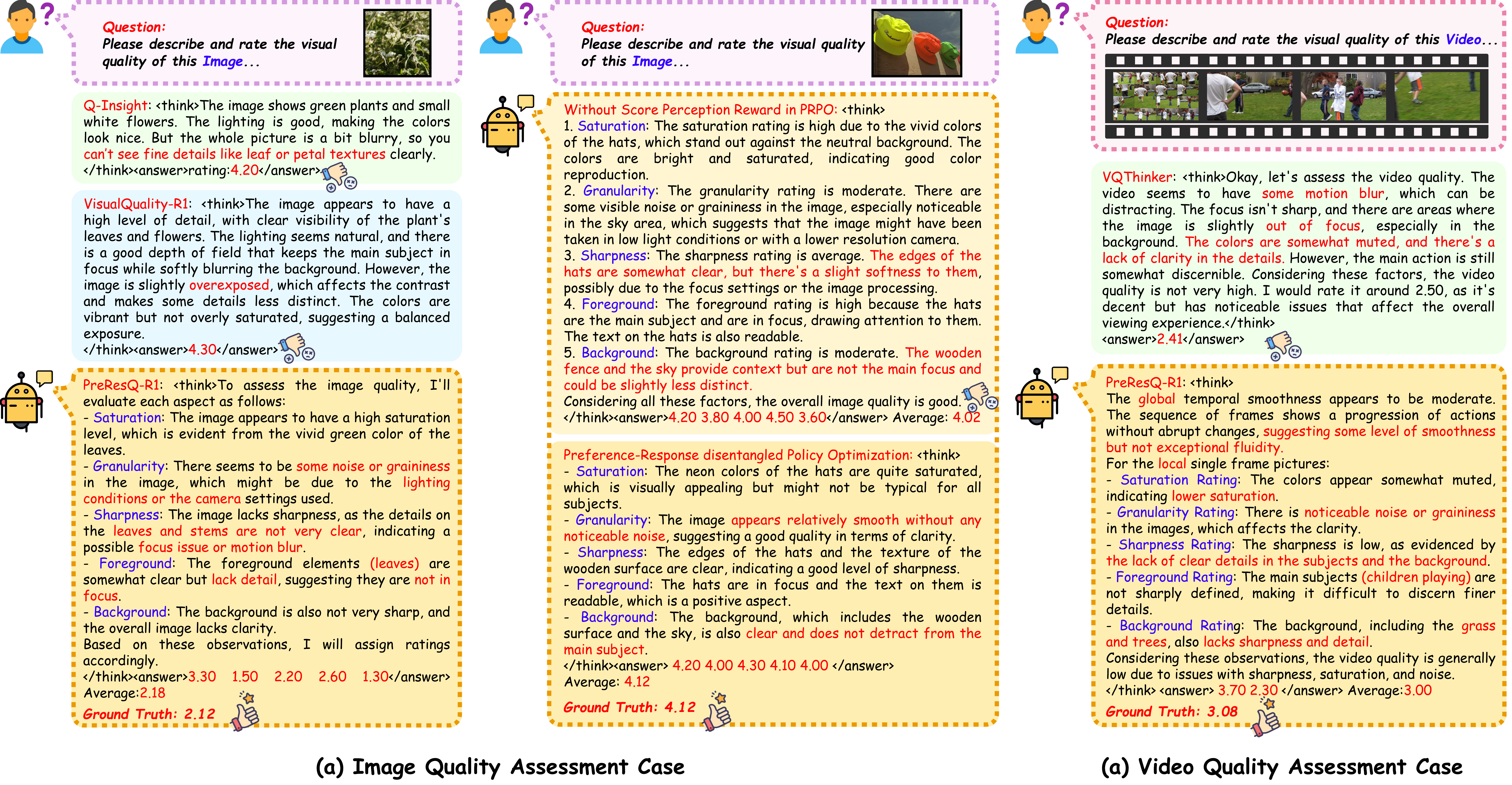}
    \vspace{-5mm}
    \caption{(a) Qualitative cases of IQA in comparison and ablation studies. (b) Qualitative cases of VQA in comparison. Fine-grained perceptual attributes are highlighted in \textcolor[RGB]{127, 0, 255}{purple}, and quality-relevant visual evidence in \textcolor{red}{red}.}
    \label{fig:5}
\vspace{-4mm}
\end{figure*}
\vspace{-2mm}
\subsection{Implementation details}
We conduct IQA PreResQ-R1 in two stages. In the first stage, prompts are randomly selected from a predefined set of $\textbf{X}=5$, with standard deviation (std) penalties set as $\delta_{\min}=0.5$ and $\lambda_{\mathrm{std}}=0.5$, and generate $K=12$ responses each sample. In the second stage, we generate $K=6$ responses each sample. We set $\alpha$=0.5, $\beta$=0.75. We fine-tune Qwen2.5-VL-7B~\cite{bai2025qwen25vltechnicalreport} with GRPO~\cite{shao2024deepseekmathpushinglimitsmathematical} without freezing any layers. For VQA task, we set $K=5$, $M=12$, and $N=3$. PreResVQA-R1 is fine-tuned based on the PreResIQA-R1 only in single-prompt stage. Training is performed using AdamW~\cite{loshchilov2019decoupledweightdecayregularization} with an initial learning rate of $3\times 10^{-7}$, a linear decay schedule. All experiments are conducted on 8 NVIDIA A800 GPUs, with a per-GPU batch size of 48 for IQA and 25 for VQA. The models are trained for 8 epochs on IQA and 1 epoch on VQA, requiring approximately 20 hours and 35 hours of total training time, respectively.
\subsection{Comparison studies}
\vspace{-2mm}
For \textbf{IQA} task, we perform quantitative and qualitative comparisons with handcrafted, deep learning-based, and MLLM-based methods, as summarized in Table~\ref{tab:comparison}. When trained on KADID-10K, PreResQ-R1 achieves an average SRCC of \textbf{0.811} and PLCC of \textbf{0.795} across 10 IQA benchmarks, surpassing prior approaches by \textbf{5.6\%} and \textbf{2.2\%}, respectively. Notably, PreResQ-R1 attains the highest SRCC on all evaluated datasets, indicating consistently improved perceptual ordering enabled by the proposed pairwise-triplet score-ranking reward. Furthermore, we also provide quantitative results trained on KonIQ dataset (Appendix \ref{sec:C}).

\textbf{Qualitative comparisons} in Fig.~\ref{fig:5}(a) show that PreResQ-R1 produces reasoning traces that are tightly grounded in visual evidence, with consistent attribution of quality-relevant factors. In contrast, existing methods such as Q-Insight and VisualQuality-R1 may produce fluent explanations that are not consistently supported by visual evidence. By explicitly constraining response consistency and preference alignment, PreResQ-R1 encourages evidence-aligned CoT reasoning, resulting in improved interpretability and reliability.

For \textbf{VQA} task, we compare PreResQ-R1 with comparable VQA methods, as reported in Table~\ref{tab:comparison_video}. PreResQ-R1 achieves an average SRCC of \textbf{0.872} and PLCC of \textbf{0.891} across 4 datasets. As shown in Fig.~\ref{fig:5}(b), compared with VQ-Thinker, PreResQ-R1 yields more accurate predictions accompanied by structured reasoning that jointly captures global temporal coherence, local spatial distortions, and fine-grained attributes.
Notably, PreResVQA-R1 is fine-tuned from PreResIQA-R1, demonstrating that perceptual priors learned from IQA can be effectively transferred to VQA. This design bridges IQA and VQA without introducing additional video-specific encoding parameters, enabling parameter-efficient adaptation while preserving strong reasoning capability under practical input constraints (Appendix \ref{text_prompt}).
\begin{table}[t]
\centering
\caption{Performance comparison of PreResQ-R1 trained on LSVQ-28K dataset and evaluated across different VQA datasets under SRCC and PLCC metrics.
\textcolor{red}{\textbf{Red}}, \textcolor{green}{\textbf{green}}, and \textcolor{blue}{\textbf{blue}} denote Handcrafted, Discriminative deep-learning, and MLLM-based methods.
\textbf{Bold} and \underline{underlined} indicate the best and second best results. $\triangleleft$ denotes trained on fused-287K dataset, $\diamond$ denotes trained on LSVQ-28K dataset, $\star$ denotes fine-tuned based on PreResIQA.}
\vspace{-3mm}
\scriptsize
\renewcommand{\arraystretch}{0.95}
\setlength{\tabcolsep}{2.3pt}
\begin{adjustbox}{width=\columnwidth}
\begin{tabular}{lcccccccccccc}
\toprule
Method &
\multicolumn{2}{c}{LSVQ$_{\text{test}}$} &
\multicolumn{2}{c}{KoNViD-1k} &
\multicolumn{2}{c}{LIVE-VQC} &
\multicolumn{2}{c}{YouTube-UGC} &
\multicolumn{2}{c}{Avg} \\
\cmidrule(lr){2-3} \cmidrule(lr){4-5} \cmidrule(lr){6-7}
\cmidrule(lr){8-9} \cmidrule(lr){10-11}
& SRCC & PLCC & SRCC & PLCC & SRCC & PLCC & SRCC & PLCC & SRCC & PLCC \\
\midrule
\rowcolor{purple!10}
STEM$\diamond$
& 0.206 & 0.243 & 0.619 & 0.627 & 0.594 & 0.629 & 0.284 & 0.318 & 0.426 & 0.454 \\
\rowcolor{purple!10}
NIQE$\diamond$
& 0.442 & 0.332 & 0.541 & 0.553 & 0.596 & 0.628 & 0.278 & 0.290 & 0.464 & 0.451 \\
\midrule
\rowcolor{green!10}
FAST-VQA$\diamond$
& 0.880 & 0.880 & 0.859 & 0.854 & \underline{0.826} & 0.845 & 0.730 & 0.747 & 0.824 & 0.832 \\
\rowcolor{green!10}
MinimalisticVQA$\diamond$
& 0.885 & 0.882 & 0.862 & 0.859 & 0.775 & 0.821 & 0.826 & 0.821 & 0.837 & 0.846 \\
\midrule
\rowcolor{blue!10}
VQ-Insight$\triangleleft$ (no code)
& 0.875 & 0.876 & 0.875 & 0.884 & 0.790 & 0.835 & -- & -- & -- & -- \\
\rowcolor{blue!10}
Q-Align$\diamond$
& \underline{0.886} & \underline{0.884}& 0.876 & 0.878 & 0.783 & 0.819 & 0.834 & 0.846 & 0.845 & 0.857 \\
\rowcolor{blue!10}
VQA$^{2}$-Scorer$\diamond$
& 0.878 & 0.872 & 0.881 & 0.880 & 0.785 & 0.830 & \textbf{0.811} & 0.823 & 0.839 & 0.851 \\
\rowcolor{blue!10}
Q-Insight$\diamond$
& 0.644 & 0.639 & 0.751 & 0.753 & 0.624 & 0.708 & 0.560 & 0.591 & 0.645 & 0.673 \\
\rowcolor{blue!10}
VisualQuality-R1$\diamond$
& 0.795 & 0.796  & 0.784 & 0.792 & 0.732 & 0.781 & 0.717 & 0.730 & 0.757 & 0.576 \\
\rowcolor{blue!10}
PreResQ-R1 (Ours)$\diamond$
& 0.865 & 0.853& 0.843 & 0.851 & 0.774 & 0.762 & 0.795 & 0.806 & 0.819 & 0.818 \\
\rowcolor{blue!10}
VQAThinker$\diamond$
& 0.883 & 0.880 & \underline{0.881} & \underline{0.884} & 0.808 & \underline{0.847} & \underline{0.860} & \underline{0.863} & \underline{0.858} & \underline{0.869} \\
\rowcolor{blue!10}
PreResQ-R1 (Ours)$\star$
& \textbf{0.901} & \textbf{0.893}& \textbf{0.879} & \textbf{0.889} & \textbf{0.838} & \textbf{0.895} & \textbf{0.870} & \textbf{0.887} & \textbf{0.872} & \textbf{0.891} \\
\bottomrule
\end{tabular}
\label{tab:comparison_video}
\end{adjustbox}
\vspace{-2mm}
\end{table}

\begin{table}[t]
\centering

\begin{minipage}{\columnwidth}
    \centering
    \captionof{table}{Performance ablation of PreResQ-R1 across different reward functions. RTR, PSR, and
    PTR represent the Response Triplet Ranking reward, Preference Pairwise Score-Ranking reward,
    and Preference Triplet Ranking reward, respectively. No performance collapse is observed under hyper-parameter variations.}
    \scriptsize
    \vspace{-2mm}
    \renewcommand{\arraystretch}{1.2}
    \setlength{\tabcolsep}{2.3pt}
    \begin{adjustbox}{width=\columnwidth}
    \begin{tabular}{ccc|cc|cc|cc|cc|cc}
    \hline
    \multirow{2}{*}{RTR} & \multirow{2}{*}{PSR} & \multirow{2}{*}{PTR} &
    \multicolumn{2}{c|}{\textbf{BID}} & \multicolumn{2}{c|}{\textbf{KonIQ}} & \multicolumn{2}{c|}{\textbf{SPAQ}} & \multicolumn{2}{c|}{\textbf{TID2013}} & \multicolumn{2}{c}{\textbf{AGIQA}} \\
    \cline{4-13}
    & & & PLCC & SRCC & PLCC & SRCC & PLCC & SRCC & PLCC & SRCC & PLCC & SRCC \\
    \hline
    \rowcolor{gray!10}\ding{55} & \checkmark & \ding{55} & 0.782 & 0.767 & 0.839 & 0.859 & 0.844 & 0.834 & 0.775 & 0.796 & 0.762 & 0.759 \\
    \rowcolor{gray!10}\ding{55} & \ding{55} & \checkmark & 0.765 & 0.790 & 0.868 & 0.875 & 0.858 & 0.829 & 0.794 & 0.781 & 0.770 & 0.764 \\
    \rowcolor{gray!10}\checkmark & \checkmark & \ding{55} & 0.795 & 0.813 & 0.872 & 0.867 & 0.882 & 0.890 & 0.823 & 0.847 & 0.791 & 0.808 \\
    \rowcolor{gray!10}\checkmark & \ding{55} & \checkmark & 0.816 & 0.809 & \textbf{0.895} & \textbf{0.904} & 0.896 & 0.889 & 0.842 & 0.858 & 0.785 & 0.779 \\
    \rowcolor{gray!10}\checkmark & \checkmark & \checkmark & \textbf{0.849} & \textbf{0.839} & 0.884 & 0.887 & \textbf{0.919} & \textbf{0.900} & \textbf{0.858} & \textbf{0.877} & \textbf{0.823} & \textbf{0.809} \\
    \hline
    \end{tabular}
    \end{adjustbox}
    \label{ablation1}
\end{minipage}

\begin{minipage}{\columnwidth}
    \centering
    \captionof{table}{Computational cost ablation of PreResQ-R1 across different reward functions (RTR, PSR, and PTR functions). TSF represents the Temporal-Spatial Flow strategy for VQA task.}
    \scriptsize
    \renewcommand{\arraystretch}{1.2}
    \setlength{\tabcolsep}{2.3pt}
    \begin{adjustbox}{width=\columnwidth}
    \begin{tabular}{cccc|cc|cc}
    \hline
    RTR & PSR & PTR & TSF &
    Training Memory & Training Time & Test Memory & Test Time\\
    \hline
    \rowcolor{gray!10}\ding{55} & \ding{55}& \checkmark&\ding{55} & 68.53$\pm$2.94 GB & 2.09$\pm$0.12 h/10K samples & 33.32$\pm$0.72 GB & 0.52 s/sample\\
    \rowcolor{gray!10}\ding{55} & \checkmark & \ding{55} & \ding{55}& 68.93$\pm$1.58 GB & 2.12$\pm$0.09 h/10K samples & 33.51$\pm$0.75 GB & 0.41 s/sample\\
    \rowcolor{gray!10}\checkmark & \checkmark & \ding{55} & \ding{55}& 70.03$\pm$1.62 GB & 2.17$\pm$0.11 h/10K samples & 33.35$\pm$0.85 GB & 0.69 s/sample\\
    \rowcolor{gray!10}\checkmark & \ding{55} & \checkmark & \ding{55}& 70.26$\pm$1.69 GB & 2.26$\pm$0.15 h/10K samples & 33.26$\pm$0.79 GB & 0.60 s/sample\\
    \rowcolor{gray!10}\checkmark & \checkmark & \checkmark&\ding{55}& 71.28$\pm$2.03 GB & 2.38$\pm$0.23 h/10K samples & 33.61$\pm$0.83 GB & 0.72 s/sample\\
    \hline
    \rowcolor{gray!30}\checkmark&\checkmark&\checkmark&\checkmark& 73.37$\pm$2.91 GB & 10.21$\pm$0.38 h/10K samples & 32.63$\pm$0.70 GB & 2.95 s/sample\\
    \rowcolor{gray!30}\multicolumn{4}{c|}{VQA-Thinker}& -- & -- & 42.62$\pm$0.61 GB & 3.83 s/sample\\
    \hline
    \end{tabular}
    \end{adjustbox}
    \label{ablation2}
\end{minipage}
\end{table}
\begin{table}[h]
\centering
\caption{Quantitative evaluation on the FineVD dataset \cite{duan2025finevqfinegrainedusergenerated} across five distortion categories (Color, Noise, Artifact, Blur, Temporal) in terms of SRCC and PLCC. Our method PreResQ consistently outperforms existing baselines including Q-Align, VisualQuality, and VQAThinker on all distortion types.}
\label{tab:finevd_performance}
\setlength{\tabcolsep}{3.2pt}
\small
\begin{adjustbox}{width=\columnwidth}
\begin{tabular}{l cc cc cc cc cc}
\toprule
\multirow{2}{*}{Model} & \multicolumn{2}{c}{Color} & \multicolumn{2}{c}{Noise} & \multicolumn{2}{c}{Artifact} & \multicolumn{2}{c}{Blur} & \multicolumn{2}{c}{Temporal} \\
\cmidrule(lr){2-3} \cmidrule(lr){4-5} \cmidrule(lr){6-7} \cmidrule(lr){8-9} \cmidrule(lr){10-11}
& SRCC & PLCC & SRCC & PLCC & SRCC & PLCC & SRCC & PLCC & SRCC & PLCC \\
\midrule
\rowcolor{gray!10}Q-Align         & 0.529 & 0.554 & 0.506 & 0.553 & 0.554 & 0.569 & 0.591 & 0.582 & 0.505 & 0.512 \\
\rowcolor{gray!10}VisualQuality   & 0.595 & 0.609 & 0.584 & 0.567 & 0.632 & 0.658 & 0.637 & 0.640 & 0.516 & 0.529 \\
\rowcolor{gray!10}VQAThinker      & 0.620 & 0.657 & 0.608 & 0.605 & 0.628 & 0.634 & 0.629 & 0.635 & 0.539 & 0.561 \\
\hline
\rowcolor{gray!30}PreResQ         & 0.655 & 0.672 & 0.635 & 0.652 & 0.664 & 0.660 & 0.665 & 0.652 & 0.595 & 0.573 \\
\bottomrule
\end{tabular}
\end{adjustbox}
\vspace{-2mm}
\end{table}

\begin{figure}[t]
    \centering
    \vspace{-2mm}
    \includegraphics[width=0.96\linewidth]{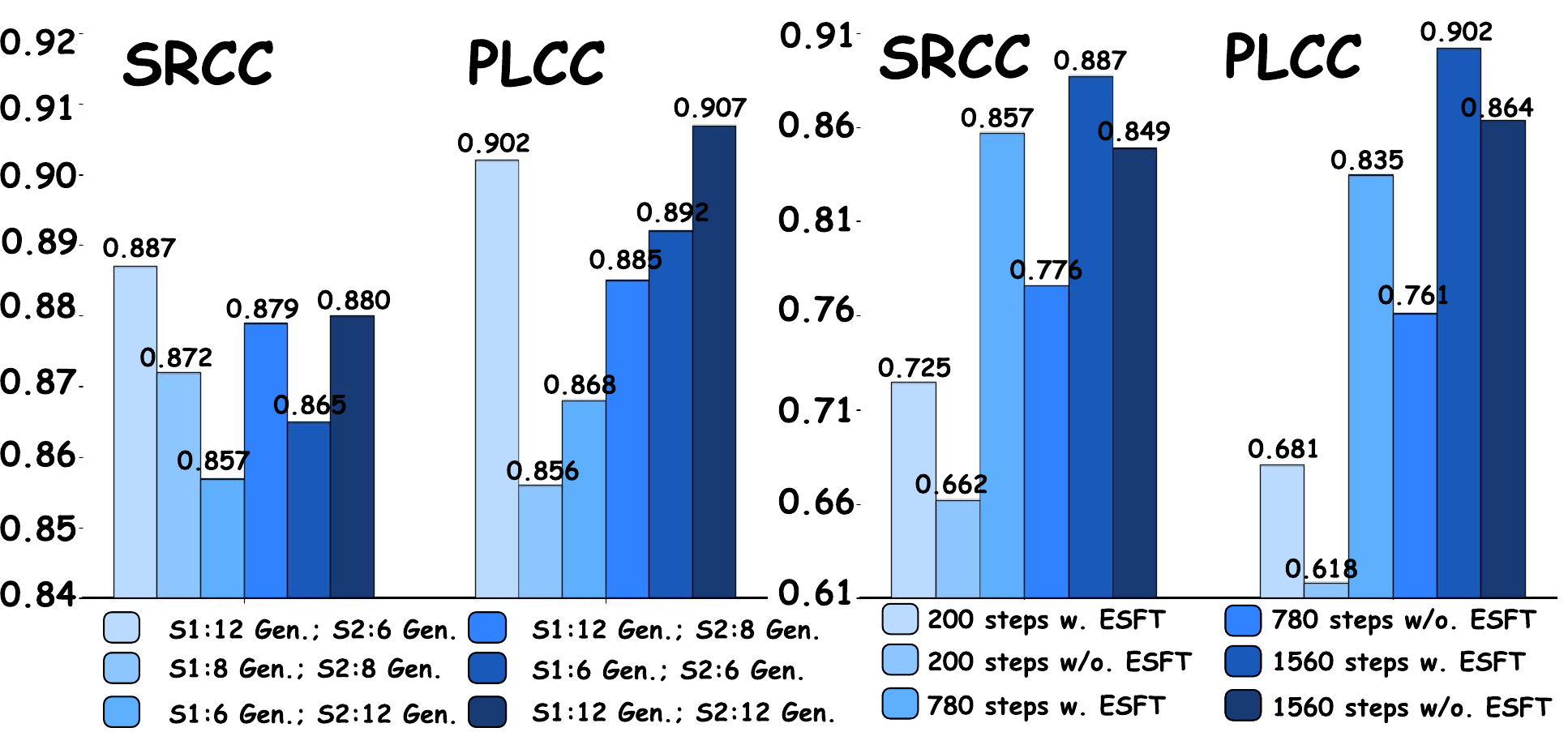}
    \vspace{-2mm}
    \caption{Left: results on 10 synthetic evaluation datasets (each dataset 100 images) with different generation number $Gen.$ across two stages $S1/S2$; right: results on evaluation datasets with/without Exploration-to-Stability Fine-Tuning (ESFT).}
    \label{fig:6}
    \vspace{-2mm}
\end{figure}
\vspace{-2mm}
\subsection{Ablation studies}
\vspace{-1mm}
We further analyze the impact of \textbf{individual reward components} in Table~\ref{ablation1}.
PSR establishes a stable baseline, confirming that pairwise preference learning is effective yet insufficient for capturing global perceptual consistency.
Incorporating RTR brings a consistent performance gain of around 3\%, highlighting the importance of constraining intra-sample response variability to stabilize reasoning and score generation.
Introducing PTR yields further improvements of further 3\% by enforcing globally higher-order perceptual ordering beyond pairwise relations. The complementary effects of Response-dim and Preference-dim (Tab.~\ref{tab:ablation_stages}) are also detailed in Appendix~\ref{sec:B}. Moreover, we provide additional \textbf{absolute-score-relative-rank ablation qualitative studies} on the IQA task in Fig.\ref{fig:5} (a) and temporal-spatial analyses on the VQA task (Appendix \ref{text_prompt}). Without the score perception reward, the reasoning CoT becomes less coherent and the score degrades significantly, demonstrating the rationality of the disentangled RL design.

Furthermore, we conduct \textbf{computational cost ablations} in Table~\ref{ablation2}.
For \textbf{IQA} task, incorporating additional reward components doesn't measurably affect training cost, with inference time at 0.6 s/sample.
For \textbf{VQA} task, the TSF strategy introduces additional cost mainly from image encoding; however, it remains substantially lighter than full video encoding, with inference time at 2.95 s/sample.
These results indicate that PreResQ-R1 doesn't introduce significant computational burden, validating its practical efficiency.

Finally, we analyze the \textbf{ablation of response generation} and the \textbf{ESFT training dynamics} on synthetic IQA dataset.
Fig.~\ref{fig:6} left shows that increasing response diversity in Stage~1 is crucial for effective exploration, while a balanced generation-comparison scheme in Stage~2 stabilizes preference learning, jointly leading to optimal performance.
As shown in Fig.~\ref{fig:6} right, ESS significantly accelerates convergence and enhances multi-granularity prediction accuracy by smoothly transitioning from exploration to stable optimization.
\subsection{Comprehensive Studies}
\vspace{-1mm}
\begin{figure}[t]
    \centering
    \vspace{-2mm}
    \includegraphics[width=0.98\linewidth]{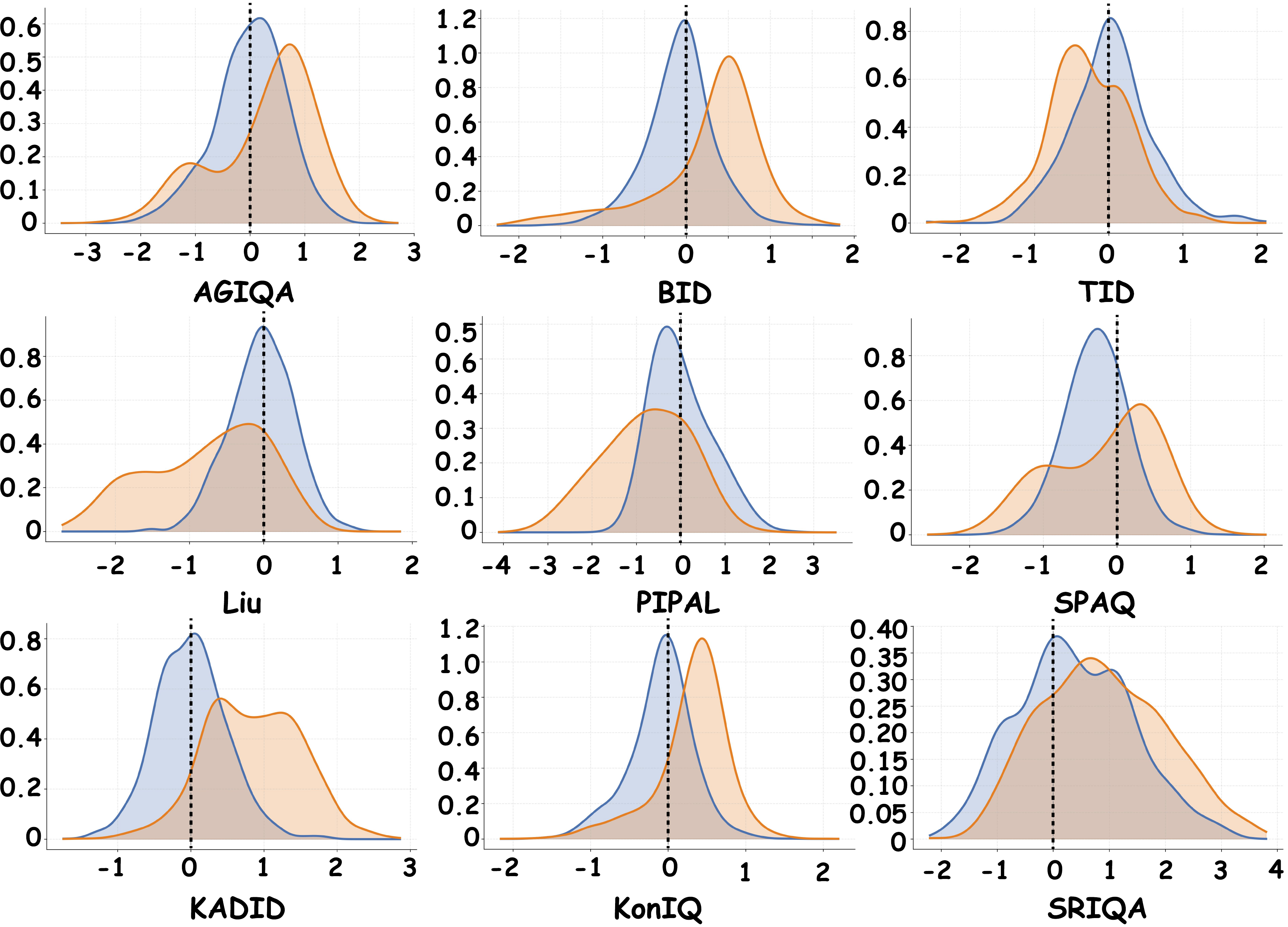}
    \caption{Comparison between PreResQ-R1 and VisualQuality-R1~\cite{wu2025visualqualityr1reasoninginducedimagequality} on distribution of difference between answer and ground truth. The horizontal axis represents the error, and the vertical axis represents the relative proportion. The closer the distribution is to 0, the better the model performance is. Blue and Orange denotes PreResQ-R1 and VisualQuality-R1.}
    \label{fig:4}
\vspace{-4mm}
\end{figure}
\begin{figure}[t]
    \centering
    \includegraphics[width=1\linewidth]{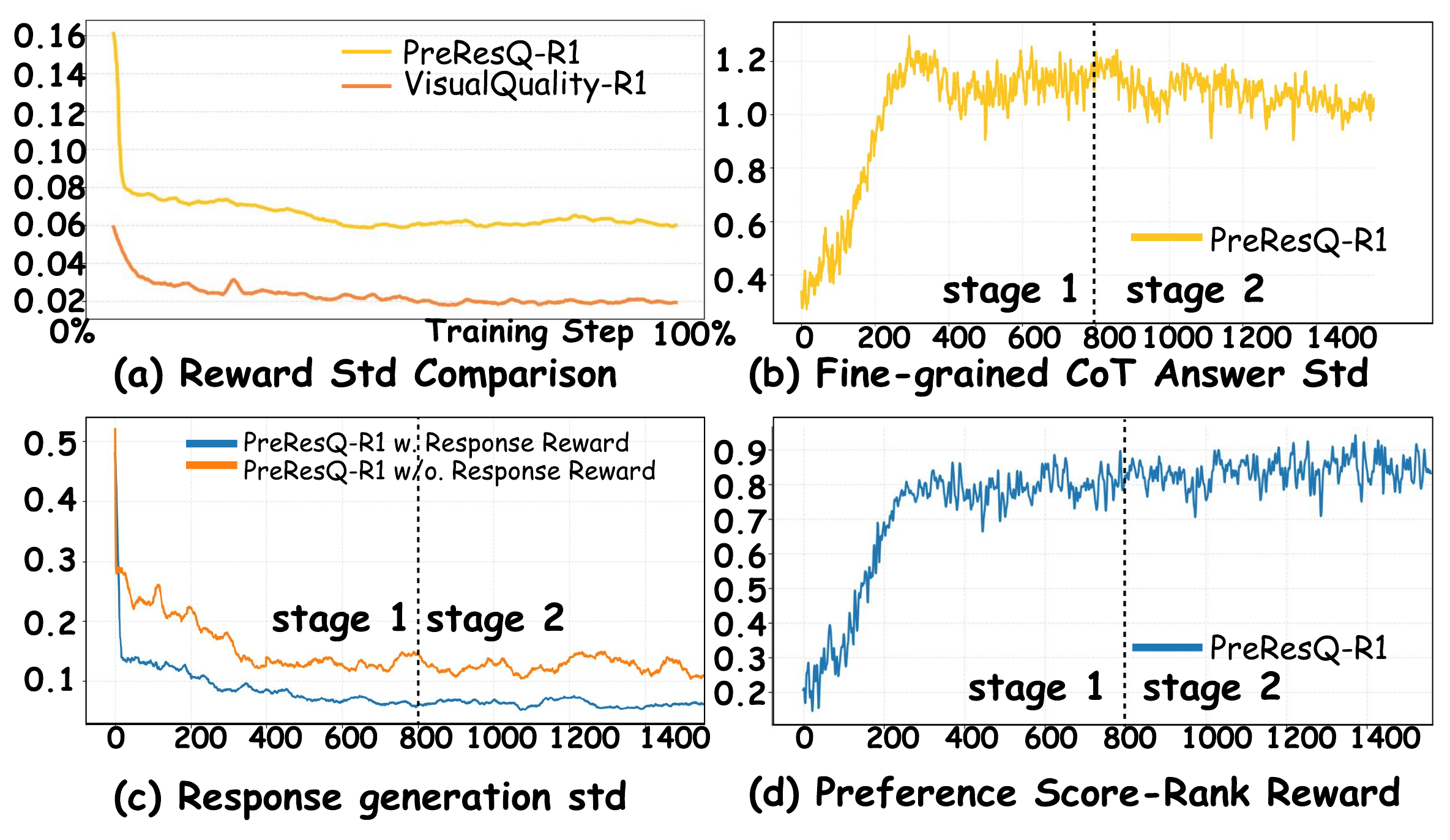}
    \vspace{-7mm}
    \caption{(a). Reward standard deviation comparison between VisualQuality-R1 and PreResQ-R1. (b). Fine-grained CoT intra answer score standard deviation. (c). generation standard deviation comparison. (d). Total reward tendency visualization.}
    \label{fig:7}
\vspace{-6mm}
\end{figure}

Fig.~\ref{fig:7} visualizes the \textbf{training dynamics} of PreResQ-R1, including reward standard deviation (std), fine-grained CoT answer std, response generation std. As shown in Fig.~\ref{fig:7}(a), the reward std demonstrates that the disentangled RL formulation moves beyond a single static target, progressively fitting discriminative pairwise and triplet preferences. Fig.~\ref{fig:7}(b) shows a low-to-high evolution of CoT standard deviation, followed by a slight reduction in Stage~2, suggesting that the std penalty functions as an early exploration scaffold and diminishes as preference-driven rewards take over with reduced policy entropy. Introducing the RTR reward further stabilizes response generation (Fig.~\ref{fig:7}(c)), constraining intra-sample variability and enabling consistent and faster convergence. 

Finally, Fig.~\ref{fig:7}(d) illustrates the progressive stabilization and improvement of the preference score-rank reward, reflecting the model’s enhanced capability to align inter-sample preferences while preserving magnitude-sensitive calibration. More hyperparamter scanning results are also available in Appendix \ref{sec:A}. 
Finally, we provide detailed evaluations of GPT-5 mini reasoning, the contributions of the PRPO and TSF modules, and a comprehensive analysis of perceptual-level performance across fine-grained IQA and VQA attributes (Appendices~\ref{sec:gpt5}, \ref{sec:tsf}, and \ref{sec:D}). These results collectively offer empirical support for the design of complementary response- and preference-level objectives, illustrating how each component contributes to improved training stability, enhanced reasoning fidelity, and consistent global perceptual alignment.
\vspace{-2mm}
\section{Limitaion and future work}
Despite its effectiveness, PreResQ-R1 has several limitations. First, the disentangled formulation assumes that response stability and preference alignment can be cleanly separated, which may not fully capture their inherent coupling in human perceptual judgments. Second, while response-level regularization improves robustness, it does not fully resolve uncertainty under highly subjective or out-of-distribution conditions, where quality perception lacks a well-defined consensus. Third, the proposed video extension relies on sparse temporal evidence and may be insufficient for scenarios requiring fine-grained motion reasoning.
Future work will explore more principled formulations of perceptual uncertainty, including unified models that jointly capture stability and preference under stochastic generation. Extending the framework to long-form and dynamic visual content with richer temporal modeling is another direction. Improving robustness under distribution shifts and incorporating adaptive or human-in-the-loop feedback may further enhance reliability.
\section{Conclusion}
\vspace{-1mm}
We introduced PreResQ-R1, a preference-response disentangled reinforcement optimization framework that unifies score regression and ranking consistency. With only 6K image samples and 28K video samples, it achieves comparable performance across 10 IQA datasets and 5 VQA datasets while providing reasonable CoT. Future work will aim to integrate image quality assessment with text-to-image generation within a unified framework, enabling complementary optimization.
\section*{Acknowledgments}
This work is partly supported by National Natural Science Foundation of China under Grant No. 72471140, partly by Shanghai Jiao Tong University Key Projects of Smart Liberal Arts under Grant No. ZHWK2505, partly by Fundamental Research Funds for the Central Universities under Grant No. YG2025QNA13, partly by Shanghai Magnolia Talent Program-Pujiang Project under Grant No. 2025PJC167, partly by the Lushan Research Funding.
\nocite{langley00}
\bibliography{example_paper}
\newpage
\appendix
\twocolumn
\section{Hyperparameter scanning results}
\label{sec:A}
\begin{figure}[t]
    \centering
    \vspace{-1mm}
    \includegraphics[width=0.96\linewidth]{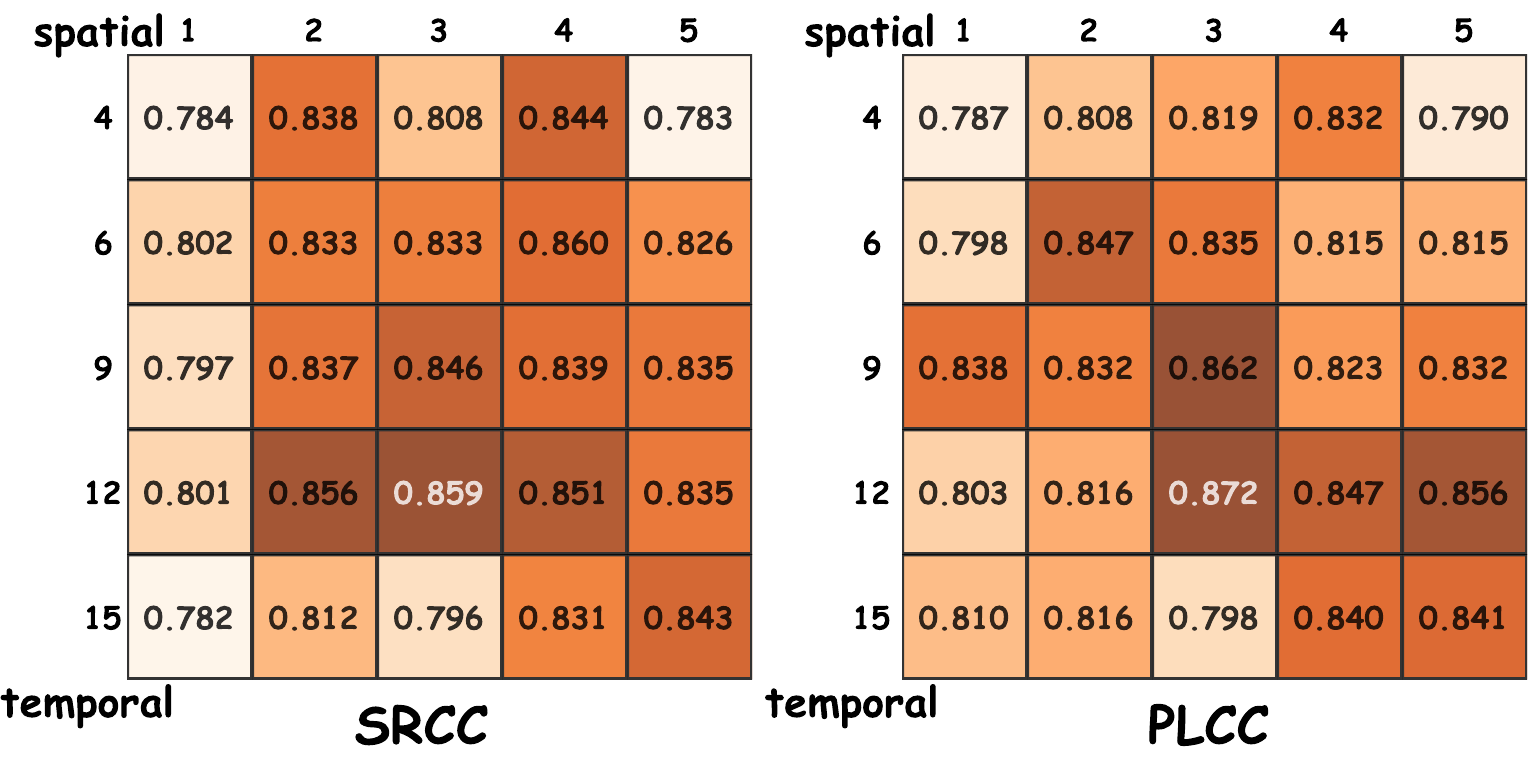}
    \vspace{-2mm}
    \caption{Performance on four synthetic VQA datasets (100 videos per dataset), peaking at temporal-flow images $M=12$ and spatial-flow images $N=3$ under hyper-parameter scanning.}
    \label{fig:8}
\end{figure}
\begin{figure}[t]
    \centering
    \vspace{-3mm}
    \includegraphics[width=0.96\linewidth]{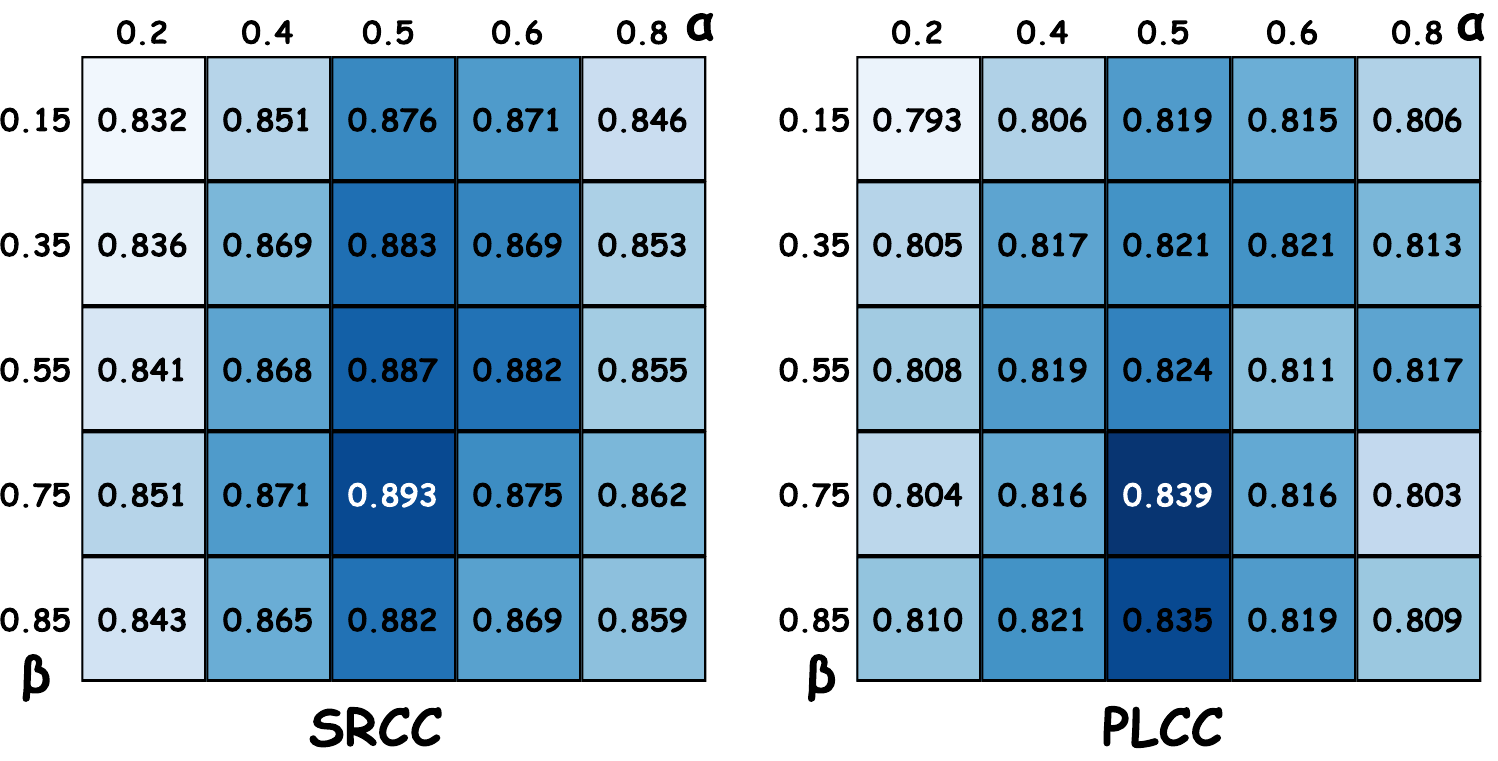}
    \caption{Performance on 10 synthetic IQA datasets (100 images per dataset) under hyper-parameter scanning of $\alpha$ and $\beta$.}
    \label{fig:65}
    \vspace{-1mm}
\end{figure}
\begin{table}[t]
  \centering
  \caption{Ablation study on different design choices of Res-dim and Pre-dim across two training stages. Results are reported with SRCC/PLCC metrics, where values in parentheses denote improvements over the baseline.}
  \label{tab:ablation_stages}
  \small
  \begin{adjustbox}{width=1\columnwidth}
  \begin{tabular}{ll cc}
    \toprule
    \textbf{Stage1} & \textbf{Stage2} & \textbf{SRCC} & \textbf{PLCC} \\
    \midrule
    Baseline & Baseline & 0.708$\pm$0.051 & 0.734$\pm$0.043 \\
    w.Res-dim & w.Res-dim & 0.721$\pm$0.006(+0.013) & 0.749$\pm$0.006(+0.015) \\
    \rowcolor{gray!20}w.Res-dim & w.Res-Pre-dim & 0.789$\pm$0.008(+0.081) & 0.783$\pm$0.007(+0.049) \\
    w.Pre-dim & w.Pre-dim & 0.754$\pm$0.042(+0.046) & 0.762$\pm$0.037(+0.028) \\
    \rowcolor{gray!20}w.Pre-dim & w.Res-Pre-dim & 0.787$\pm$0.010(+0.079) & 0.780$\pm$0.009(+0.046) \\
    \rowcolor{gray!20}w.Res-Pre-dim & w.Res-Pre-dim & 0.811$\pm$0.008(+0.103) & 0.790$\pm$0.007(+0.056) \\
    w.Res-Pre-dim & w.Res-dim & 0.809$\pm$0.015(+0.101) & 0.785$\pm$0.014(+0.051) \\
    \bottomrule
  \end{tabular}
  \end{adjustbox}
\end{table}
\begin{table*}[t]
\centering
\caption{Performance of PreResQ-R1 on KonIQ datasets across SRCC and PLCC. \textcolor{red}{\textbf{Red}}, \textcolor{green}{\textbf{green}}, and \textcolor{blue}{\textbf{blue}} denote handcrafted, discriminative deep-learning, and MLLM-based methods. \textbf{Bold} and \underline{underline} denote the best and second best results. All results are obtained by averaging over ten independent runs with different random seeds.}
\scriptsize
\renewcommand{\arraystretch}{0.95}
\setlength{\tabcolsep}{2.2pt}
\begin{adjustbox}{width=2.1\columnwidth}
\begin{tabular}{l cc cc cc cc cc cc cc cc cc cc cc}
\toprule
{\textbf{Method}}
& \multicolumn{2}{c}{\textbf{SPAQ}}
& \multicolumn{2}{c}{\textbf{BID}}
& \multicolumn{2}{c}{\textbf{CLIVE}}
& \multicolumn{2}{c}{\textbf{Liu13}}
& \multicolumn{2}{c}{\textbf{SRIQA}}
& \multicolumn{2}{c}{\textbf{TID13}}
& \multicolumn{2}{c}{\textbf{KonIQ}}
& \multicolumn{2}{c}{\textbf{KADID}}
& \multicolumn{2}{c}{\textbf{PIPAL}}
& \multicolumn{2}{c}{\textbf{AQIOA}}
& \multicolumn{2}{c}{\textbf{Avg}} \\
\cmidrule(lr){2-23}
& \textbf{SRCC} & \textbf{PLCC}
& \textbf{SRCC} & \textbf{PLCC}
& \textbf{SRCC} & \textbf{PLCC}
& \textbf{SRCC} & \textbf{PLCC}
& \textbf{SRCC} & \textbf{PLCC}
& \textbf{SRCC} & \textbf{PLCC}
& \textbf{SRCC} & \textbf{PLCC}
& \textbf{SRCC} & \textbf{PLCC}
& \textbf{SRCC} & \textbf{PLCC}
& \textbf{SRCC} & \textbf{PLCC}
& \textbf{SRCC} & \textbf{PLCC} \\
\midrule

\rowcolor{purple!10}BRISQUE
& 0.406 & 0.490 & 0.329 & 0.337 & 0.256 & 0.267 & 0.352 & 0.347 & 0.187 & 0.149 & 0.531 & 0.550 & 0.226 & 0.225 & 0.356 & 0.429 & 0.232 & 0.267 & 0.497 & 0.541 & 0.337 & 0.360 \\

\rowcolor{purple!10}NIQE
& 0.664 & 0.679 & 0.394 & 0.405 & 0.309 & 0.329 & 0.309 & 0.317 & 0.215 & 0.190 & 0.509 & 0.521 & 0.533 & 0.530 & 0.405 & 0.468 & 0.161 & 0.195 & 0.533 & 0.560 & 0.403 & 0.419 \\

\rowcolor{green!10}MUSIQ
& 0.863 & 0.868 & 0.516 & 0.513 & 0.287 & 0.312 & 0.347 & 0.360 & 0.254 & 0.227 & 0.627 & 0.634 & 0.929 & 0.924 & 0.556 & 0.575 & 0.431 & 0.431 & 0.630 & 0.720 & 0.544 & 0.557 \\

\rowcolor{green!10}ManIQA
& 0.768 & 0.864 & 0.492 & 0.510 & 0.291 & 0.305 & 0.494 & 0.485 & 0.234 & 0.225 & 0.681 & 0.673 & 0.758 & 0.768 & 0.499 & 0.654 & 0.457 & 0.419 & 0.723 & 0.685 & 0.540 & 0.559 \\

\rowcolor{green!10}CLIP-IQA+
& 0.866 & 0.758 & 0.527 & 0.536 & 0.356 & 0.391 & 0.515 & 0.524 & 0.459 & 0.439 & 0.643 & 0.652 & 0.864 & 0.866 & 0.653 & 0.465 & 0.427 & 0.452 & 0.736 & 0.636 & 0.605 & 0.572 \\

\rowcolor{blue!10}Qwen2.5-VL
& 0.869 & 0.872 & 0.711 & 0.725 & 0.604 & 0.635 & 0.742 & 0.738 & 0.441 & 0.418 & 0.702 & 0.682 & 0.899 & 0.889 & 0.675 & 0.693 & 0.384 & 0.375 & 0.752 & 0.785 & 0.677 & 0.681 \\

\rowcolor{blue!10}DeQA
& 0.896 & 0.895 & 0.716 & 0.714 & 0.616 & 0.662 & 0.761 & 0.749 & 0.510 & 0.504 & 0.659 & 0.661 & \underline{0.941} & \underline{0.953} & 0.687 & 0.694 & 0.478 & 0.472 & 0.815 & 0.814 & 0.708 & 0.712 \\

\rowcolor{blue!10}C2Score
& 0.860 & 0.867 & 0.725 & 0.729 & 0.627 & 0.657 & 0.775 & 0.765 & 0.431 & 0.409 & 0.726 & 0.751 & 0.881 & 0.875 & 0.687 & 0.694 & 0.342 & 0.345 & 0.729 & 0.809 & 0.678 & 0.690 \\

\rowcolor{blue!10}Q-Align
& 0.887 & 0.886 & 0.707 & 0.719 & 0.694 & 0.713 & 0.805 & 0.791 & 0.479 & 0.428 & 0.754 & 0.769 & 0.940 & 0.941 & 0.684 & 0.674 & 0.419 & 0.403 & 0.735 & 0.772 & 0.710 & 0.710 \\

\rowcolor{blue!10}Q-Insight
& \underline{0.905} & \textbf{0.907} & \underline{0.755} & \underline{0.759} & 0.715 & 0.751 & \textbf{0.815} & \underline{0.806} & \underline{0.561} & \textbf{0.550} & 0.729 & 0.732 & 0.916 & 0.933 & \underline{0.736} & \underline{0.742} & \underline{0.486} & \underline{0.474} & 0.764 & 0.811 & 0.738 & \underline{0.747} \\

\rowcolor{blue!10}VisualQuality-R1
& 0.892 & 0.891 & 0.743 & 0.751 & \underline{0.749} & \underline{0.776} & 0.784 & 0.785 & 0.554 & 0.531 & \underline{0.769} & \underline{0.775} & 0.915 & 0.924 & 0.711 & 0.712 & 0.438 & 0.441 & \underline{0.867} & \underline{0.822} & \underline{0.742} & 0.741 \\

\rowcolor{blue!10}PreResQ-R1 (Ours)
& \textbf{0.919} & \underline{0.904} & \textbf{0.761} & \textbf{0.765} & \textbf{0.762} & \textbf{0.792} & \underline{0.810} & \textbf{0.808} & \textbf{0.579} & \underline{0.541} & \textbf{0.774} & \textbf{0.781} & \textbf{0.951} & \textbf{0.959} & \textbf{0.741} & \textbf{0.750} & \textbf{0.491} & \textbf{0.482} & \textbf{0.882} & \textbf{0.835} & \textbf{0.767} & \textbf{0.762} \\

\bottomrule
\end{tabular}
\end{adjustbox}
\label{tab:comparison_single}  
\end{table*}
For \textbf{VQA} task, a \textbf{hyperparameter sweep} over $M$ and $N$ in Table~\ref{fig:8}) shows that performance increases with more global-temporal frames, peaking at $M=12$ due to downsampling trade-offs. In contrast, once multiple local-spatial frames are jointly aggregated for reasoning ($N>2$), performance stabilizes, highlighting the robustness of PreResIQA-R1 as the base for local evidence integration.

For \textbf{IQA} task, we conduct a \textbf{hyperparameter sweep} over $\alpha$ and $\beta$ in Table~\ref{fig:65}.
Both SRCC and PLCC remain stable across a broad range of values, indicating low sensitivity to hyperparameter choices, with peak performance achieved at $\alpha\!=\!0.5$ and $\beta\!=\!0.75$.
Beyond scalar metrics, the error distributions in Fig.~\ref{fig:4} show that PreResQ-R1 produces tighter and more centered predictions, offering stronger and more reliable evidence of its superiority in quality score estimation.
\section{Ablation experiment on reward}
\label{sec:B}
Tab. \ref{tab:ablation_stages} further reveals the complementary, non-redundant nature of Response-dim and Preference-dim: gray-highlighted configurations integrating both dimensions outperform single-dimension setups, confirming joint optimization of response stability and preference alignment as the performance key. The optimal configuration (Res-Pre-dim in both stages) achieves SRCC 0.811±0.008 (+0.103) and PLCC 0.790±0.007 (+0.056), supporting our insight that response stabilization enables more effective preference learning.

Quantitatively, Response-dim and Preference-dim contribute 37\% and 50\% to performance gains, respectively: Preference-dim enhances global human preference alignment, while Response-dim suppresses intra-sample variability, stabilizes optimization, and boosts generalization. Critical for MLLM-based QA under limited supervision (6K images, 28K videos), Response-dim ensures stable performance across 10 benchmarks. These results confirm the rationality of PRPO and the necessity of joint response-preference optimization to address trajectory instability and distribution mismatch overlooked by existing methods.
\section{Evaluation of Reasoning Quality}
\label{sec:gpt5}
As shown in Table~\ref{tab:gpt5mini_evaluation}, PreRes-IQA and PreRes-VQA consistently outperform their respective baselines across all four GPT-5 mini reasoning dimensions: Accuracy, Reasoning, Completeness, and Confidence. This indicates that the model generates structured and coherent CoT outputs that capture meaningful perceptual cues beyond simple score prediction.
\section{Temporal-Spatial Flow Effectiveness}
\label{sec:tsf}
In addition, the ablation results in Table~\ref{tab:ablation_module} show substantial improvements in SRCC and PLCC when incorporating the Preference-Response Disentangled Policy Optimization and Temporal-Spatial Flow modules. This confirms that TSF effectively aggregates global and local visual evidence for video quality assessment, while PRPO stabilizes intra-sample responses and aligns inter-sample preferences.
\section{Performance on KonIQ dataset}
\label{sec:C}
\begin{table}[t]
  \centering
  \caption{Quantitative evaluation on the GPT-5 mini benchmark over four reasoning dimensions: Accuracy, Reasoning, Completeness, and Confidence. Higher scores denote better performance.}
  \label{tab:gpt5mini_evaluation}
  \small
  \begin{adjustbox}{width=1\columnwidth}
  \begin{tabular}{l cccc}
    \toprule
    \textbf{Model} & \textbf{Accuracy} & \textbf{Reason} & \textbf{Completeness} & \textbf{Confidence} \\
    \midrule
    VisualQuality-R1 & 3.94 & 4.69 & 4.71 & 4.66 \\
    \rowcolor{gray!20} PreRes-IQA       & \textbf{4.65} & \textbf{5.21} & \textbf{5.85} & \textbf{5.15} \\
    \midrule
    VQAThinker       & 3.62 & 3.94 & 3.82 & 4.15 \\
    \rowcolor{gray!20} PreRes-VQA       & \textbf{3.94} & \textbf{4.71} & \textbf{4.62} & \textbf{4.59} \\
    \bottomrule
  \end{tabular}
  \end{adjustbox}
\end{table}
\begin{table}[t]
  \centering
  \caption{Ablation study on the effectiveness of proposed Preference-Response disentangled Policy Optimization (PRPO) and Temporal-Spatial Flow (TSF) modules. Values in parentheses indicate the improvements over the baseline.}
  \label{tab:ablation_module}
  \small
  \begin{adjustbox}{width=0.55\columnwidth}
  \begin{tabular}{l cc}
    \toprule
    \textbf{Model} & \textbf{SRCC} & \textbf{PLCC} \\
    \midrule
    \textbf{Baseline} & 0.767 & 0.621 \\
    \midrule
    \textbf{w. PRPO} & 0.819(+53\%) & 0.793(+68\%) \\
    \midrule
    \rowcolor{gray!20}\textbf{w. TSF} & 0.865(+47\%) & 0.872(+32\%) \\
    \bottomrule
  \end{tabular}
  \end{adjustbox}
  \vspace{-4mm}
\end{table}

\begin{table}[t]
  \centering
  \caption{Fine-grained CoT performance on SRCC metric in IQA task.}
  \label{tab:supp_f}
  \small
  \begin{adjustbox}{width=1\columnwidth}
  \begin{tabular}{l|ccccc|c}
    \toprule
    \textbf{Task} & \textbf{Saturation} & \textbf{Granularity} & \textbf{Sharpness} & \textbf{Foreground} & \textbf{Background} & \textbf{SRCC} \\
    \midrule
    IQA  & 0.613 & 0.431 & 0.470 & 0.564 & 0.264 & 0.811 \\
    \bottomrule
  \end{tabular}
  \end{adjustbox}
\end{table}
\begin{table}[t]
  \centering
  \caption{Global-local CoT performance on SRCC metric in VQA task.}
  \label{tab:supp_g}
  \small
  \begin{adjustbox}{width=0.5\columnwidth}
  \begin{tabular}{l|cc|c}
    \toprule
    \textbf{Task} & \textbf{Global} & \textbf{Local} & \textbf{SRCC} \\
    \midrule
    VQA  & 0.619 & 0.284 & 0.865 \\
    \bottomrule
  \end{tabular}
  \end{adjustbox}
  \vspace{-4mm}
\end{table}
We further assess cross-dataset generalization to evaluate the robustness and transferability of PreResQ-R1. As reported in Table~\ref{tab:comparison_single}, when trained solely on KonIQ, PreResQ-R1 achieves an average SRCC of 0.758 and PLCC of 0.743 across four distinct IQA datasets, corresponding to improvements of 12.70\% and 7.09\% over the Qwen2.5-VL baseline. These results indicate that the preference-response disentangled reinforcement learning framework not only stabilizes intra-sample reasoning but also facilitates the learning of perceptual representations that generalize effectively across diverse visual domains. The substantial gains in cross-dataset performance highlight the model’s ability to disentangle and robustly encode attribute-level quality cues, suggesting that the learned representations capture fundamental perceptual principles rather than overfitting to dataset-specific score distributions.
\section{Perceptual-Level Performance Analysis}
\label{sec:D}
The results in Tables \ref{tab:supp_f} and \ref{tab:supp_g} reveal nuanced insights into PreResQ-R1’s handling of perceptual dimensions across IQA and VQA tasks. For IQA, while the overall SRCC reaches 0.811, the contribution of individual attributes varies considerably: saturation (0.613) and foreground quality (0.564) align strongly with human judgments, whereas background fidelity (0.264) and granularity (0.431) show weaker correlations. This divergence indicates that PreResQ-R1 captures heterogeneous, attribute-specific visual cues that are not trivially aggregated into the final score, reflecting the model’s ability to maintain discriminative intermediate representations and disentangle fine-grained perceptual factors from global assessments. In VQA, the global-temporal context exhibits higher predictive strength (SRCC 0.619) than local-spatial cues (0.284), yet their combination reaches 0.865, highlighting the model’s capacity to integrate complementary evidence across temporal and spatial scales. These findings suggest that robust video quality assessment benefits not from over-reliance on either temporal coherence or frame-level detail, but from structured aggregation that preserves the informational distinctiveness of each dimension. It demonstrates that PRPO not only stabilizes global predictions but also cultivates interpretable, multi-dimensional perceptual representations.
\section{Further methodological insights}
\begin{figure*}[t]
    \centering
    \includegraphics[width=0.6\linewidth]{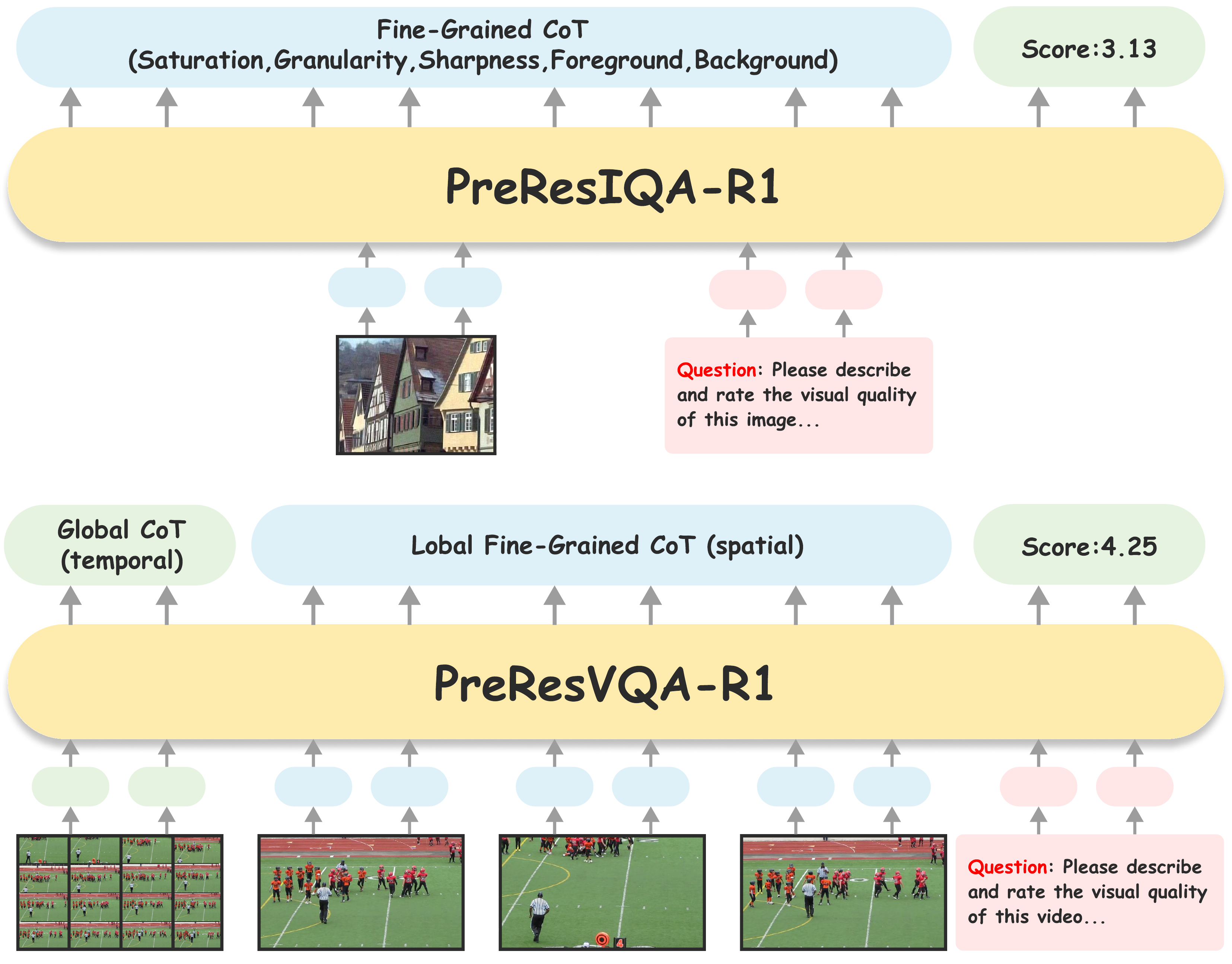}
    \caption{The input and output data flow in PreResIQA-R1 and PreResVQA-R1.}
    \label{fig:IQA_VQA}
    \vspace{-2mm}
\end{figure*}
\textbf{Spatiotemporal aggregation in VQA: global-temporal and local-spatial decoupling.} Inspired by DeepSeek-OCR~\cite{wei2025deepseekocrcontextsopticalcompression}, which compresses textual sequences into compact image tokens, we extend this principle to video quality assessment by decoupling temporal and spatial evidence. The global-temporal branch captures overall motion consistency and smoothness by compressing multiple frames into a single multiframe representation, enabling efficient reasoning over temporal coherence. The local-spatial branch evaluates perceptual quality within individual frames, sampling three key frames to balance accuracy and computational cost. Unlike prior methods such as MinimalisticVQA~\cite{sun2024analysisvideoqualitydatasets} and VQAThinker~\cite{cao2025vqathinkerexploringgeneralizableexplainable}, which rely on downsampling or reinforcement for temporal consistency, our approach explicitly disentangles temporal dynamics from spatial fidelity. Pretraining on image-level quality provides strong initialization for video-level assessment, demonstrating that decoupled aggregation combined with cross-modal transfer yields robust and interpretable video quality evaluation.\\
\textbf{Why we divide visual assessment into fine-grained details.} Human perception of visual quality is intrinsically multi-dimensional, encompassing interrelated factors such as texture fidelity, color saturation, contrast, and semantic clarity. Conventional IQA models that regress a single holistic score fail to capture these heterogeneous perceptual cues, leading to entangled representations and unstable reasoning. By decomposing assessment into five fine-grained perceptual dimensions (Saturation, Granularity, Sharpness, Foreground, and Background) PreResQ-R1 explicitly structures the visual reasoning process. This decomposition enforces a dimension-aware reasoning pathway that aligns linguistic explanations with visual evidence, thereby promoting fine-grained perceptual reasoning. The structured formulation further stabilizes optimization through dimension-wise regularization (e.g., standard deviation penalty) and reward balancing, ultimately enhancing interpretability, robustness, and perceptual fidelity.\\
\textbf{Why we introduce a preference-response disentangled policy optimization definition.} Conventional reward formulations in RL-based IQA reduce perceptual alignment to either scalar regression or ordinal ranking, neglecting the intertwined nature of intra-sample reasoning stability and inter-sample perceptual consistency that underpin human quality judgment. The proposed Preference-Response Disentangled Policy Optimization establishes a dual-axis reward geometry that decomposes optimization into response coherence and preference alignment. The former regularizes reasoning trajectories within each perceptual context, while the latter constrains global quality ordering across heterogeneous samples. This disentangled formulation transforms reward learning from distribution-specific scalar fitting into a structurally grounded perceptual reasoning process, preventing spurious reward exploitation and overfitting to dataset priors. By explicitly balancing local response fidelity and global perceptual regression, PRPO enforces reward discrimination, enhances policy robustness against distributional shift.\\
\textbf{Why we train only on KADID-10K dataset, and evaluate PreResQ-R1 on ten datasets.} Training solely on KADID-10K deliberately targets preference alignment rather than re-teaching visual perception: we assume the underlying MLLM already possesses strong visual understanding and reasoning priors, and \textbf{our goal is to use minimal RL fine-tuning to correct its deviations from human aesthetic preference}. This low-cost RL calibration nudges the model’s pre-trained perceptual priors toward human MOS with limited supervision. Evaluation across ten heterogeneous benchmarks that cover both synthetic and authentic distortions constitutes a rigorous out-of-distribution test. The consistently high SRCC and PLCC values illustrate that our disentangled PRPO framework effectively mitigates human-model preference bias while retaining the inherent generalization ability of the MLLM. In short, this work emphasizes efficient alignment (minimal RL intervention) over wholesale perceptual re-learning, offering a scalable path for human-aligned aesthetic assessment.\\
\textbf{Why we voluntarily introduce std penalty and response balance reward to obtain Exploration-to-Stability fine-tuning process.} We introduce a standard deviation penalty at the fine-grained answer level in first exploration stage to accelerate the model’s sensitivity to differences across granular aspects, enabling it to accurately predict each fine-grained component. The response balance reward promotes consistent and stable generations, which not only enhances overall training stability but also supports reliable preference ranking and score-based rewards. These mechanisms ensure that the model learns nuanced distinctions while maintaining coherent and calibratable outputs.\\
\textbf{Why we introduce the preference triplet ranking score with the introduction of pairwise ranking and score reward.} While pairwise comparisons capture local perceptual consistency, they are insufficient to enforce global transitivity across multiple samples. The preference triplet ranking-score reward introduces a challenging and fine-grained relational constraint by modeling third-order transitivity, guiding the policy to refine subtle perceptual distinctions and construct globally coherent hierarchies. By integrating pairwise and triplet consistency, we form a synergistic reward landscape in which pairwise terms anchor local human alignment, while triplet terms eliminate cyclic inconsistencies and enhance global ranking stability. \\
\textbf{Multi-prompts in exploration-fine-tuning stage: why it works.} In the early exploration phase, fixed prompts tend to restrict reasoning diversity and confine the model’s attention to narrow perceptual patterns. Introducing multiple semantically varied prompts (e.g., focusing on “texture clarity,” “color naturalness,” or “composition balance”) stimulates broader CoT trajectories and exposes the model to different linguistic formulations of quality reasoning. This multi-prompt strategy acts as reward-level data augmentation, allowing the policy to explore a richer reasoning space and uncover diverse perceptual cues before convergence. Subsequent stability refinement distills this diversity into a calibrated policy, improving both reasoning coherence and generalization to unseen distortions.\\
\textbf{Two-stage exploration-stability fine-tuning's design considerations.} The “exploration-stability” fine-tuning paradigm shares substantial similarities with DeepSeek-R1 \cite{deepseekai2025deepseekr1incentivizingreasoningcapability}, which combines multi-prompt exploration with stability-driven reinforcement learning refinement. In general, early exploration allows the model to probe diverse reasoning trajectories and perceptual cues, while the stability phase consolidates these experiences into a consistent policy. However, without clear evidence of modifications or novel mechanisms beyond prior two-stage frameworks, it is difficult to attribute unique methodological advances to this design choice. Careful justification and ablation are therefore needed to substantiate the novelty and effectiveness of such two-stage fine-tuning.
\section{Prompt templates and qualitative cases}
\label{text_prompt}
\begin{figure*}[t]
    \centering
    \includegraphics[width=1\linewidth]{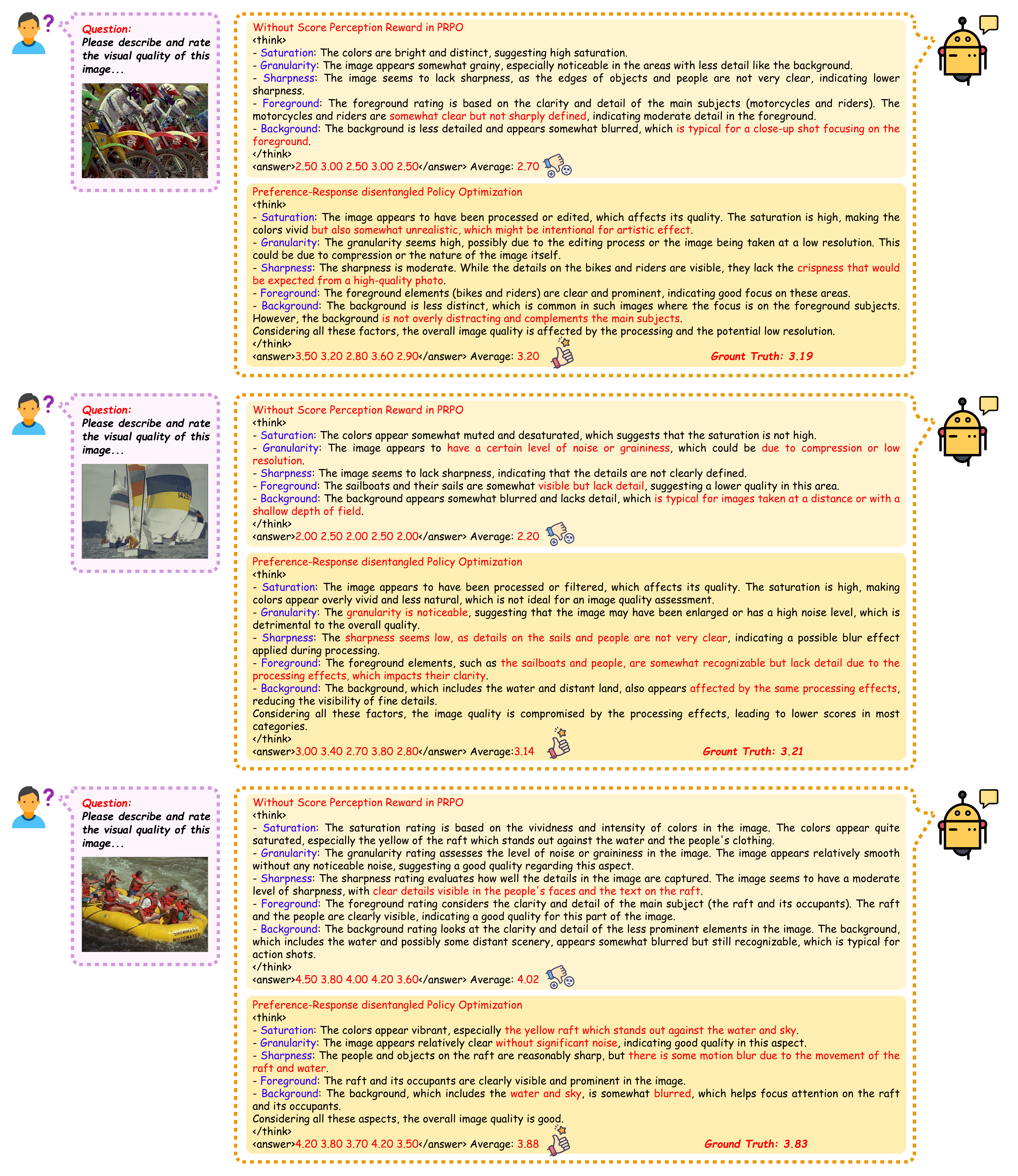}
    \vspace{-3mm}
    \caption{Qualitative ablation cases of PreResQ-R1's image quality assessment results with/without Score Perception Reward in Preference-Response disentangled Policy Optimization.}
    \label{fig:sup3}
\vspace{-5mm}
\end{figure*}

\begin{figure*}[t]
    \centering
    \includegraphics[width=1\linewidth]{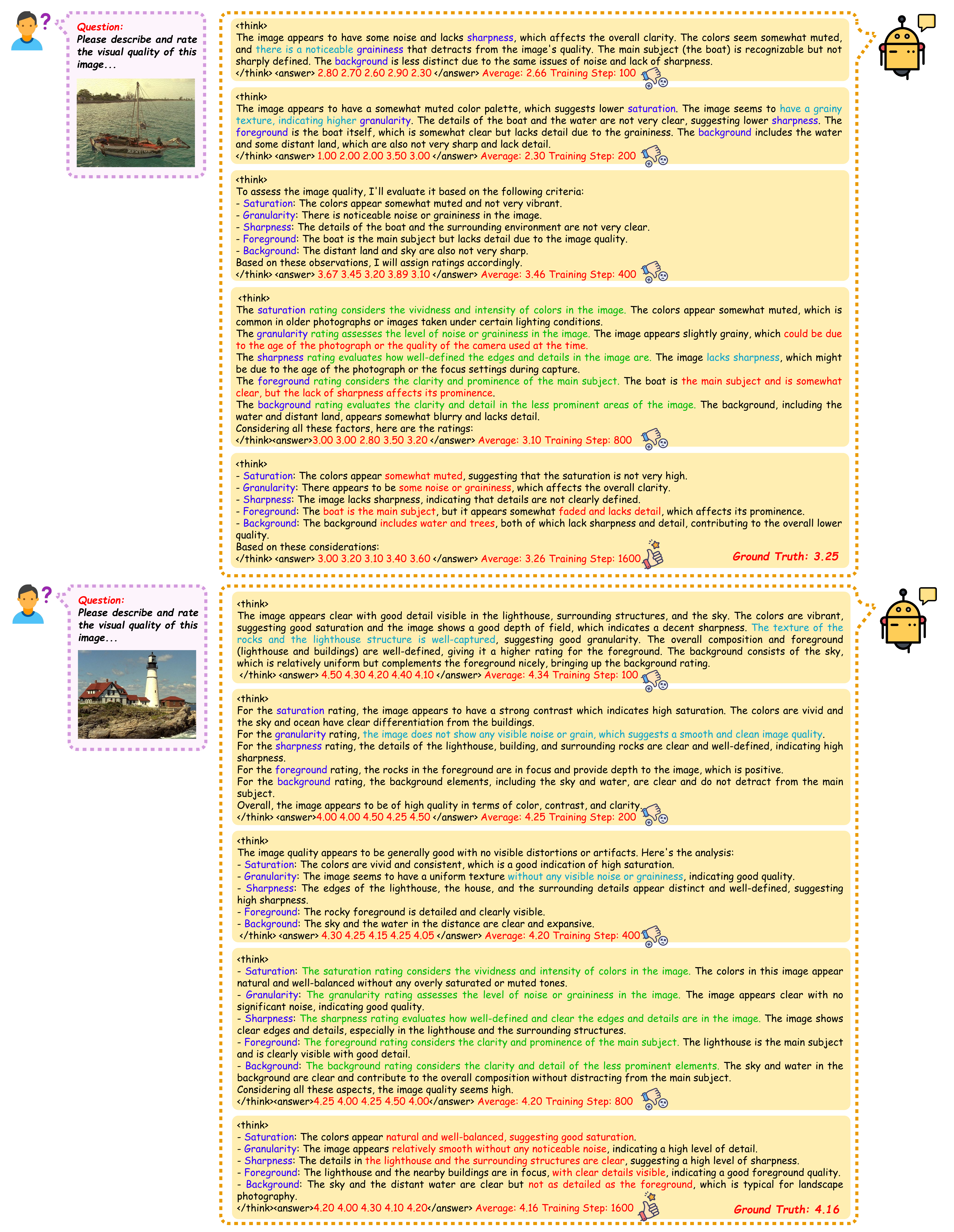}
    \vspace{-6mm}
    \caption{Qualitative comparison cases of PreResQ-R1's image quality assessment results across different training steps.}
    \label{fig:sup2}
\vspace{-5mm}
\end{figure*}

\begin{figure*}[t]
    \centering
    \includegraphics[width=1\linewidth]{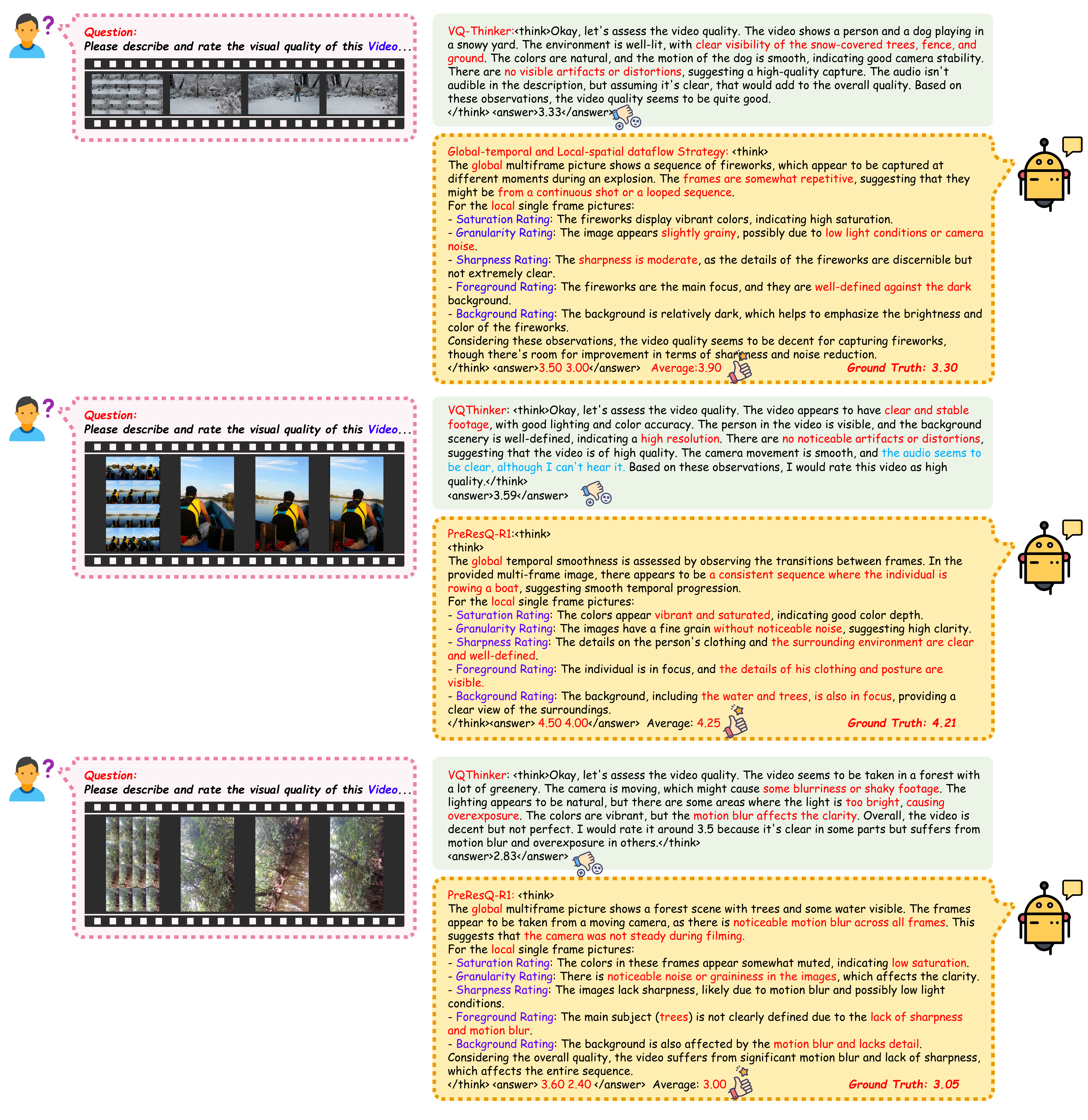}
    \vspace{-3mm}
    \caption{Qualitative comparison cases of video quality assessment results between PreResQ-R1 and state-of-the-art reinforcement learning based reasoning MLLM. Given an video sample, and a text prompt, our model performs quality assessment across saturation, granularity, sharpness, foreground, and background through CoT reasoning.}
    \label{fig:sup4}
\vspace{-5mm}
\end{figure*}

\begin{figure*}[t]
    \centering
    \includegraphics[width=1\linewidth]{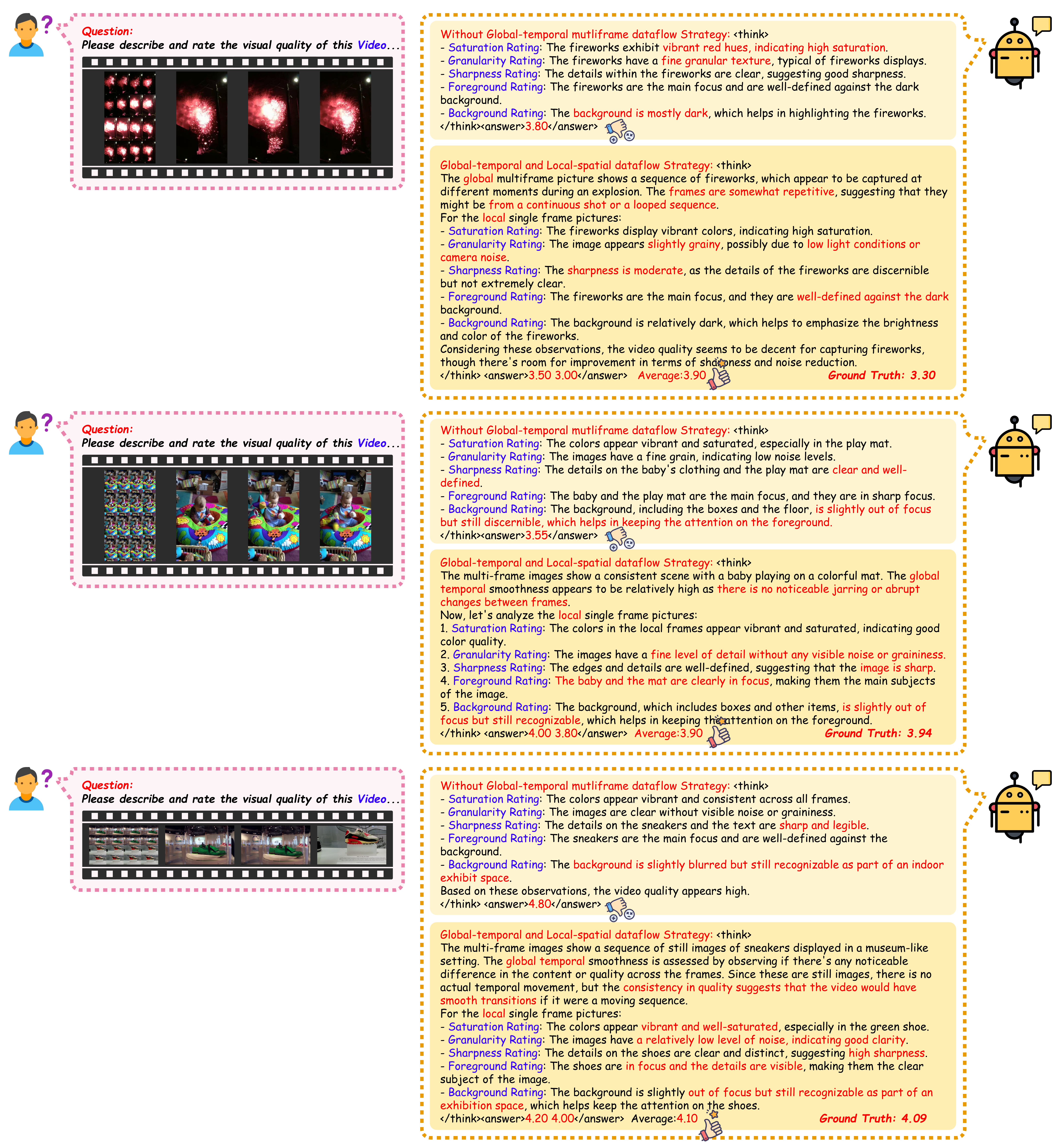}
    \vspace{-3mm}
    \caption{Qualitative ablation cases of PreResQ-R1's video quality assessment results with/without Global-Temporal dimension CoT in reasoning process.}
    \label{fig:sup5}
\vspace{-5mm}
\end{figure*}

\begin{figure*}[t]
    \centering
    \includegraphics[width=1\linewidth]{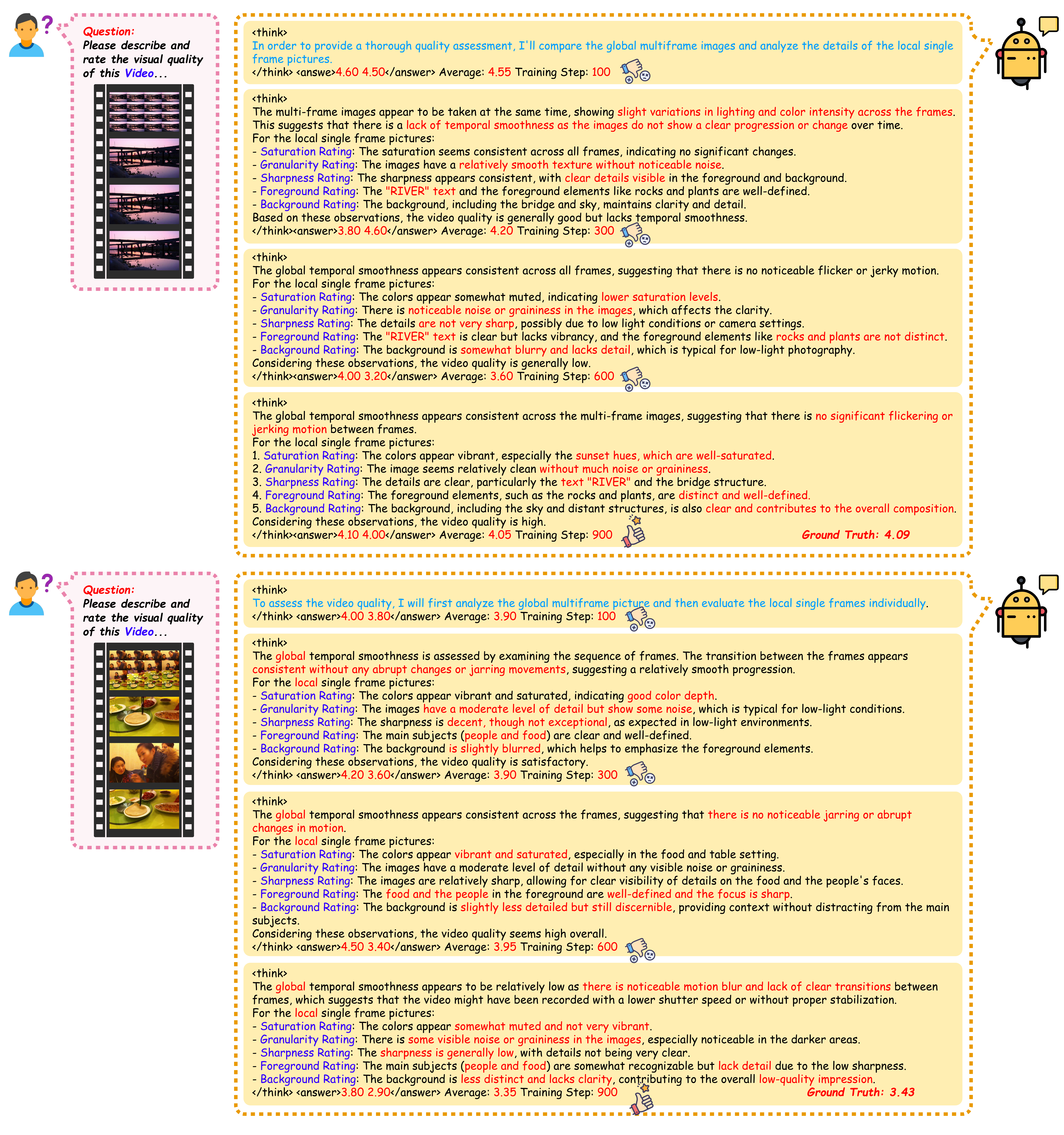}
    \vspace{-6mm}
    \caption{Qualitative comparison cases of PreResQ-R1's video quality assessment results across different training steps.}
    \label{fig:sup6}
\vspace{-5mm}
\end{figure*}

We provide a detailed overview of the overall algorithmic workflow in Algorithm \ref{alg:preresiqa}, and present the complete prompt templates used for both IQA and VQA tasks in Table~\ref{tab:prompts}. To further complement our main results, we include extensive qualitative analyses and ablation studies in the supplementary figures. Specifically, Fig.~\ref{fig:sup1}, Fig.~\ref{fig:sup2}, and Fig.~\ref{fig:sup3} illustrate additional comparison cases with state-of-the-art methods, as well as preference-response ablation analyses and step-wise training behaviors for the IQA task. Similarly, Fig.~\ref{fig:sup4}, Fig.~\ref{fig:sup5}, and Fig.~\ref{fig:sup6} present corresponding results for the VQA task, offering a more comprehensive understanding of our model’s reasoning ability, robustness, and cross-domain generalization.
\onecolumn
\begin{figure*}[t]
    \centering
    \includegraphics[width=1\linewidth]{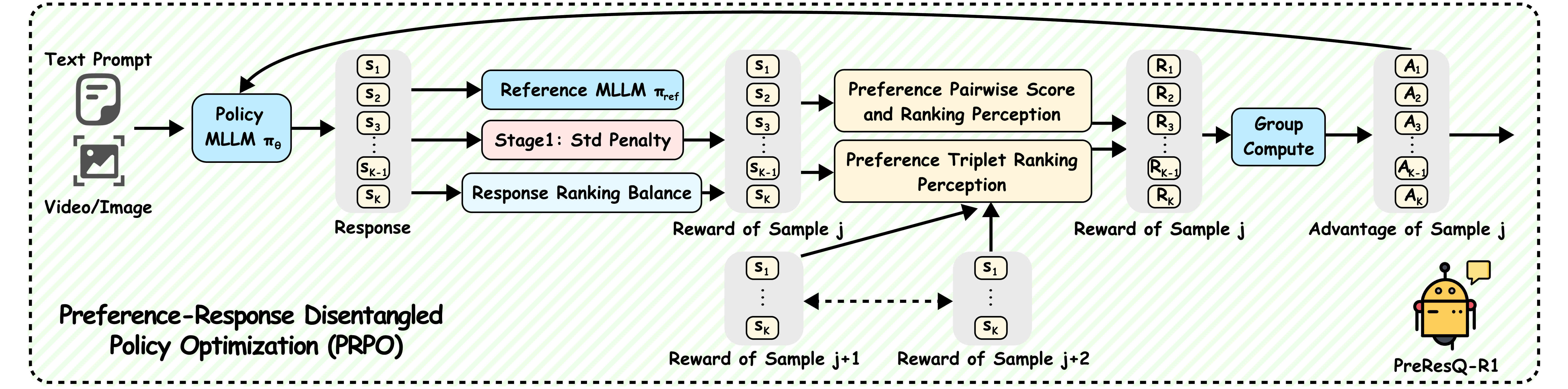}
    \vspace{-4mm}
    \caption{Pipeline of the Preference-Response Disentangled Policy Optimization (PRPO), which introduced response-based ranking reward, preference-based pairwise score-ranking reward, and preference-based triplet ranking reward to optimize group policy learning.}
    \label{fig:3}
\vspace{-2mm}
\end{figure*}
\section{Reinforcement learning optimization}
\label{app:rl}
We adopt a regularized policy optimization framework to fine-tune the MLLM using the proposed reward. For each input sample $j$, we generate $K$ stochastic trajectories and collect their total rewards into a vector:
\begin{equation}
\mathbf{r}(s_j)
= [R_{\mathrm{total}}^{j,1}, \dots, R_{\mathrm{total}}^{j,K}]^\top.
\end{equation}
To reduce variance across stochastic generations, we compute a normalized relative advantage:
\begin{equation}
a(s_j^{(i)})
= \frac{R_{\mathrm{total}}^{j,i} - \mu(\mathbf{r}(s_j))}
{\sigma(\mathbf{r}(s_j))},
\end{equation}
where $\mu(\cdot)$ and $\sigma(\cdot)$ denote the mean and standard deviation over the $K$ trajectories of sample $j$.
We optimize the policy $\pi_\theta$ using a clipped surrogate objective with KL regularization to a fixed reference policy $\pi_{\mathrm{ref}}$, following standard practice:
\begin{equation}
\mathcal{L}(\theta)
=\frac{1}{B K} \sum_{j=1}^{B} \sum_{i=1}^{K}
\min \Bigg[
\frac{\pi_\theta(o_i|c,s_j)}{\pi_\theta^{\mathrm{old}}(o_i|c,s_j)} a(s_j^{(i)}), \mathrm{clip}\!\Big(
\frac{\pi_\theta(o_i|c,s_j)}{\pi_\theta^{\mathrm{old}}(o_i|c,s_j)},
1-\epsilon, 1+\epsilon
\Big) a(s_j^{(i)})
\Bigg]
- \lambda \, D_{\mathrm{KL}}\!\left(
\pi_\theta \,\|\, \pi_{\mathrm{ref}}
\right),
\end{equation}
where $\epsilon$ is the clipping threshold and $\lambda$ controls the strength of KL regularization.
The KL divergence is approximated as:
\begin{equation}
D_{\mathrm{KL}}(\pi_\theta \,\|\, \pi_{\mathrm{ref}})
\approx
\pi_{\mathrm{ref}} \frac{\pi_\theta}{\pi_{\mathrm{ref}}}
- \log \frac{\pi_\theta}{\pi_{\mathrm{ref}}} - 1,
\end{equation}
which constrains the updated policy to remain close to the pretrained MLLM.
Finally, policy optimization follows the standard policy gradient formulation:
\begin{equation}
\nabla_{\theta} J(\theta)
=
\mathbb{E}_{\tau \sim \pi_\theta}
\Big[
R_{\mathrm{total}}
\sum_{t=1}^{T}
\nabla_\theta \log \pi_\theta(a_t|s_t)
\Big],
\end{equation}
where $a_t$ denotes the generated token at step $t$ and
$s_t = (\mathcal{P}, \mathcal{I}, a_{<t})$ is the corresponding state.

\section{Ethical considerations}
This work aims to improve the reliability of visual quality assessment in multimodal systems, but several risks remain. First, QA models reflect subjective human preferences, and biases in training data may propagate into systematic disparities across content or demographic groups. Second, increased scoring stability may obscure underlying uncertainty, potentially leading to overconfident judgments in sensitive applications. Third, large multimodal models incur nontrivial computational and environmental costs. We stress that PreResQ-R1 is intended as a research framework rather than a definitive metric, and requires careful validation before deployment in real-world settings.
\begin{algorithm*}[t]
\caption{\textbf{PreResQ-R1: Two-stage fine-grained rank-and-score reinforcement learning for image quality assessment via preference-response disentangled policy optimization.}}
\label{alg:preresiqa}
\begin{spacing}{1.05}
\begin{algorithmic}[1]
\Require Training set $\mathcal{D}=\{(\mathcal{I}_j,\mathcal{P}_j,\mathbf{G}_j)\}_{j=1}^{B}$, pretrained policy $\pi_{\theta}$, reference policy $\pi_{\mathrm{ref}}$,  
generation number $K$, batch size $B$, total steps $T$, reward weights $\alpha,\beta_{\mathrm{err}},\beta_{\mathrm{rank}}$,  
and standard deviation penalty parameters $\lambda_{\mathrm{std}},\Omega_{\mathrm{std}}$.
\Ensure Updated policy $\pi_{\theta}$.
\vspace{2pt}
\State \textbf{Stage 1: Multi-prompt Exploration with Std Penalty}
\For{each image or video sample $\mathcal{I}_i$ in dataset}
    \State Randomly select a prompt $\mathcal{P}_x$ from predefined set $\{\mathrm{P_I},...,\mathrm{P_V}\}$.
    \State Generate $K$ responses $\tau_j^{(1:K)} \sim \pi_{\theta}(\cdot|\mathcal{P}_x,\mathcal{I}_j)$.
    \State Extract dimension-wise scores $\mathbf{s}_j^{(i)}=[s_j^1,...,s_j^5]$.
    \State Compute std penalty $R_{\mathrm{std}}(\mathbf{s}_j^{(i)})$ as Eq.~(16).
    \State Compute preference-response disentangled rewards 
    $R_{\text{for}}, R_{\text{loc}}^{j,i}, R_{\text{pair}}^{j,i}, R_{\text{tri}}^{j,i}, R_{\mathrm{std}}$ (Eq.~(2-9)).
    \State Form total reward $R_j^{(i)}$ as Eq.~(10-11).
    \State Update $\pi_{\theta}$ via PRPO objective (Eq.~(12-15)).
\EndFor
\State $\triangleright$ \textit{Exploration RL2RS fine-tuning with PRPO optimization.}
\vspace{2pt}
\State \textbf{Stage 2: Single-prompt Stability Refinement}
\For{each image or video sample $\mathcal{I}_i$ in dataset}
    \State Use single standard prompt $\mathcal{P}$ for generation.
    \State Generate $K$ responses $\tau_j^{(1:K)} \sim \pi_{\theta}(\cdot|\mathcal{P},\mathcal{I}_j)$.
    \State Extract predicted scores $\mathbf{s}_j^{(i)}$.
    \State Compute rewards $R_{\text{for}}, R_{\text{loc}}^{i,k}, R_{\text{pair}}^{i,k}, R_{\text{tri}}^{i,k}$ (Eq.~(2-9)).
    \State Obtain final reward $R_j^{(i)}$ and normalize advantages $a_j^{(i)}$ (Eq.~(10-11)).
    \State Update $\pi_{\theta}$ via PRPO objective (Eq.~(12-15)).
\EndFor
\State $\triangleright$ \textit{Stability RL2RS fine-tuning with PRPO optimization.}

\State \Return Optimized policy $\pi_{\theta}$.
\end{algorithmic}
\end{spacing}
\end{algorithm*}
\begin{table*}[htbp]
\centering
\caption{Fine-grained structured text prompts for IQA and VQA tasks.}
\label{tab:prompts}
\renewcommand{\arraystretch}{1.3}
\begin{tabular}{p{0.97\linewidth}}
\toprule
\textbf{IQA Prompt 1} \\
\midrule
You are doing an image quality assessment. Provide five ratings in the following order: saturation, granularity, sharpness, foreground, and background. Please rate 5 scores. Each rating should be a float between 1 and 5, rounded to two decimals (1 = very poor, 5 = excellent). \\
\midrule
\textbf{IQA Prompt 2} \\
\midrule
You are doing an image quality assessment. For this image quality assessment task, please evaluate along five aspects: color saturation, texture granularity, clarity or sharpness, foreground quality, and background quality. Please rate 5 scores each row. Give each aspect a score between 1.00 and 5.00 (two decimals, where 1 = very poor and 5 = excellent). \\
\midrule
\textbf{IQA Prompt 3} \\
\midrule
You are doing an image quality assessment. Please assess image quality by providing five ratings in sequence: rate the saturation, rate the granularity, rate the sharpness, rate the foreground quality, and rate the background quality. Please rate 5 scores. All answers must be floats from 1.00 to 5.00, rounded to two decimals. \\
\midrule
\textbf{IQA Prompt 4} \\
\midrule
You are doing an image quality assessment. Provide ratings for five dimensions: saturation to reflect color vividness, granularity to capture texture smoothness, sharpness to evaluate detail clarity, foreground quality to assess object visibility, and background quality to measure scene consistency. Please rate 5 scores each row. Each score must be a float between 1.00 and 5.00, where 1 means poor and 5 means excellent. \\
\midrule
\textbf{IQA Prompt 5} \\
\midrule
You are doing an image quality assessment. Please give an image quality evaluation with five scores in the following order: saturation, granularity, sharpness, foreground quality, and background quality. Please rate 5 scores each row. Each score should be a float between 1 and 5, rounded to two decimals, with 1 indicating very poor quality and 5 indicating excellent quality. \\
\midrule
\textbf{IQA Template} \\
\midrule
You are doing the image quality assessment task. Please describe and rate the picture in five parts: saturation rating; granularity rating; sharpness rating; foreground rating; and background rating. All ratings should be floats between 1 and 5, rounded to two decimals, where 1 represents very poor quality and 5 represents excellent quality. \\
$<$think$>...\text{Reasoning CoT}...</$think$>$  
$<$answer$>$\text{Score1; Score2; Score3; Score4; Score5}$</$answer$>$ \\
\midrule
\textbf{VQA Template}\\
\midrule
You are doing the video quality assessment task. Please compare the internal packed video frames in the first picture and describe the global multiframe pictures in temporal deminsion and global scene. Then compare and describe the rest local single frame pictures in five parts: saturation rating; granularity rating; sharpness rating; foreground rating; and background rating. Finally, please rate two scores in the global multiframe and local single frame dimension. All ratings should be floats between 1 and 5, where 1 represents very poor quality and 5 represents excellent quality.\\
$<$think$>...\text{Reasoning CoT}...</$think$>$  
$<$answer$>$\text{Score1; Score2}$</$answer$>$ \\
\bottomrule
\end{tabular}
\end{table*}

\begin{figure*}[t]
    \centering
    \includegraphics[width=1\linewidth]{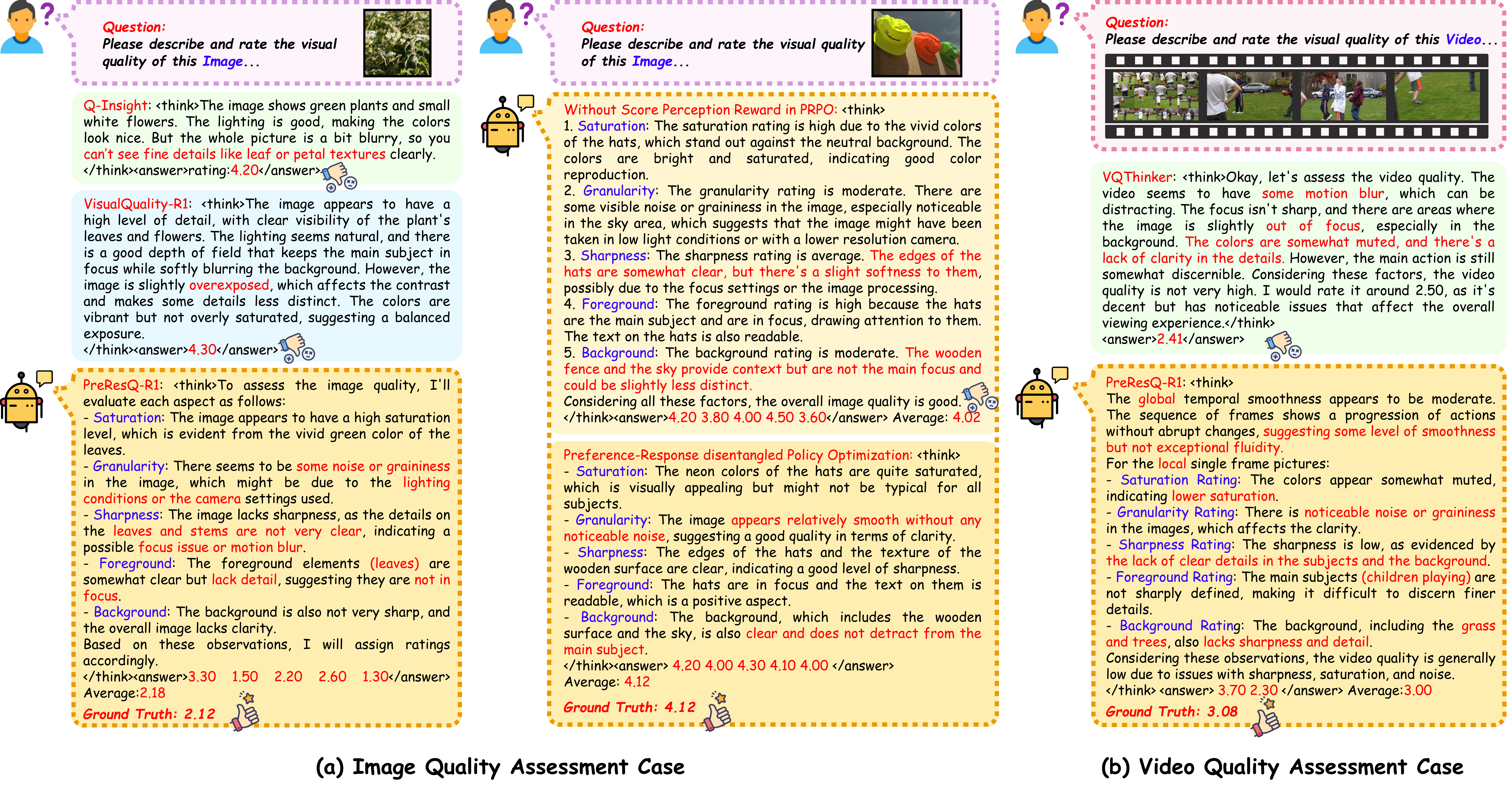}
    \vspace{-3mm}
    \caption{Qualitative comparison cases of image quality assessment results between PreResQ-R1 and state-of-the-art reinforcement learning based reasoning MLLM. Given an image, and a text prompt, our model performs quality assessment across saturation, granularity, sharpness, foreground, and background through CoT reasoning.}
    \label{fig:sup1}
\vspace{-5mm}
\end{figure*}
\twocolumn
\section{Compared Methods}
\begin{enumerate}
    \item \textbf{NIQE}~\cite{6353522}: a handcrafted no-reference IQA model that assesses image quality by measuring deviations from natural scene statistics, without relying on any human-rated training data or prior knowledge of distortions.

    \item \textbf{STEM}~\cite{9633248}: a training-free method for assessing the quality of user-generated videos. Inspired by the human visual system’s perceptual straightening hypothesis, STEM measures how distortions increase the curvature of a video’s perceptual trajectory.
    
    \item \textbf{BRISQUE}~\cite{6272356}: a handcrafted NR-IQA model that evaluates image quality by measuring deviations from natural scene statistics of locally normalized luminance in the spatial domain, without requiring distortion-specific features or reference images.
    
    \item \textbf{UNIQUE}~\cite{Zhang_2021}: a deep learning-based BIQA model that learns to assess image quality through pairwise comparison and fidelity loss, enabling unified training across both synthetic and real-world distortions for robust generalization.
    
    \item \textbf{MUSIQ}~\cite{9710973}: a deep learning-based IQA model that leverages a multi-scale Transformer architecture to assess image quality directly from images of arbitrary resolutions, capturing quality information at multiple granularities without the need for resizing or cropping.
    
    \item \textbf{MANIQA}~\cite{yang2022maniqamultidimensionattentionnetwork}: a deep learning-based NR-IQA model that employs a multi-dimensional attention architecture combining channel- and spatial-attention mechanisms (via Transposed Attention Block and Scale Swin Transformer Block) to jointly capture global and local visual dependencies, achieving superior performance—especially on GAN-distorted images.
    
    \item \textbf{LIQE}~\cite{zhang2023blindimagequalityassessment}: a MLMM-based BIQA method that leverages a multitask learning framework for joint learning of image quality assessment, scene classification, and distortion identification. By aligning visual and textual embeddings via cosine similarity, the model enables automatic parameter sharing and loss weighting, leading to more robust and generalizable quality prediction.

    \item \textbf{FAST-VQA}~\cite{wu2022fastvqaefficientendtoendvideo}: is an efficient deep video quality assessment framework that introduces a Grid Mini-patch Sampling strategy to retain both local and global quality information while drastically reducing computational cost. By processing these quality-preserving fragments through a specialized Fragment Attention Network, FAST-VQA enables effective end-to-end learning of video-quality-related representations.
    \item \textbf{VQA$^2$-Scorer}~\cite{jia2024vqa2visualquestionanswering}: is a VQA model in the VQA2 series that leverages large multi-modal model architectures to predict quantitative quality scores for videos. It processes both spatial and temporal information by interleaving visual and motion tokens, enabling detailed perception of distortions and artifacts. Built upon the VQA2 Instruction Dataset, it is trained to capture nuanced quality differences across diverse video types, achieving strong performance in both traditional quality scoring and holistic visual quality understanding tasks.
    \item \textbf{VQ-Insight}~\cite{zhang2025vqinsightteachingvlmsaigenerated}: is a reasoning-driven MLLM framework designed for evaluating the quality of AI-generated videos (AIGC). It introduces a progressive video quality learning paradigm: starting from image-level warm-up, advancing to temporal understanding, and finally achieving joint optimization with video generators. Moreover, it incorporates multi-dimensional scoring, preference comparison, and temporal modeling rewards to balance generalization and specialization. Extensive experiments show that VQ-Insight outperforms state-of-the-art methods across multiple evaluation settings, offering reliable and interpretable video quality assessment for AIGC applications.
    \item \textbf{VQThinker}~\cite{cao2025vqathinkerexploringgeneralizableexplainable}: is a reasoning-based VQA framework that integrates LMMs with reinforcement learning to jointly perform video quality understanding and scoring, mimicking human perceptual judgment. It introduces a GRPO algorithm that incorporates three VQA-specific reward mechanisms, including bell-shaped regression, pairwise ranking, and temporal consistency, to enhance reasoning quality and prediction accuracy. Experiments show that VQAThinker achieves state-of-the-art performance, strong out-of-distribution generalization, and superior explainability in both quality scoring and distortion interpretation tasks.
    \item \textbf{Q-Align}~\cite{wu2023qalignteachinglmmsvisual} : a MLLM-based method that trains LMMs for image and video quality assessment by emulating human raters’ use of discrete text-defined rating levels, enabling unified evaluation across IQA, IAA, and VQA tasks while achieving state-of-the-art performance.
    
    \item \textbf{MinimalisticVQA}~\cite{sun2024analysisvideoqualitydatasets}: a ultra-simple blind video quality assessment model designed to analyze and benchmark VQA datasets. They are built using only basic components: a video preprocessor (for spatiotemporal downsampling), a spatial quality analyzer, an optional temporal quality analyzer, and a quality regressor, all implemented with the simplest possible methods. Despite their simplicity, MinimalisticVQA models reveal fundamental issues in existing VQA datasets, such as the easy dataset problem, and provide insights for designing more robust and generalizable BVQA models.
    
    \item \textbf{DeQA-Score}~\cite{you2025teachinglargelanguagemodels}: a MLLM-based method that leverages MLLMs for precise image quality score regression by discretizing continuous score distributions into soft labels, preserving inter-image relationships, and employing a Thurstone-based fidelity loss to co-train across multiple IQA datasets, achieving stable performance aligned with human annotations.
    
    \item \textbf{Q-Insight}~\cite{li2025qinsightunderstandingimagequality}: an MLLM-based approach that leverages reinforcement learning via GRPO to jointly optimize score regression and degradation perception, enabling visual reasoning for image quality understanding under limited supervision. However, its reward design primarily focuses on absolute score regression, which often leads to overfitting specific distributions and poor robustness when generalized to unseen datasets.
    
    \item \textbf{VisualQuality-R1}~\cite{wu2025visualqualityr1reasoninginducedimagequality}: an MLLM-based reasoning-driven no-reference image quality assessment model that employs reinforcement learning to rank for human-aligned, context-aware evaluation. Its pairwise ranking formulation emphasizes relative ordering without sufficient score calibration, resulting in score distributions that deviate from human perceptual preferences.
    \item \textbf{Qwen2.5-VL-7B}~\cite{bai2025qwen25vltechnicalreport}: is a pre-trained MLLM baseline used for comparison, providing strong multi-modal representation capabilities without task-specific fine-tuning.
    \item \textbf{UnifiedReward-Think}~\cite{wang2025unifiedmultimodalchainofthoughtreward}: a unified multimodal reward model that extends UnifiedReward by incorporating explicit long CoT reasoning across both visual understanding and generation tasks. It employs a three-stage optimization process, including CoT reward initialization, rejection sampling-based generalization fine-tuning, and GRPO reinforcement refinement. By leveraging both explicit and implicit CoT reasoning, it achieves robust alignment with human perceptual judgments across diverse tasks.
    \item \textbf{Our PreResQ-R1}: a Preference-Response Disentangled Reinforcement Learning framework for fine-grained and interpretable image quality assessment. It unifies absolute score regression and relative ranking consistency within a single reasoning-driven optimization paradigm. Specifically, PreResQ-R1 decomposes the reward into two complementary branches: a response-based component that enhances intra-sample coherence across multiple reasoning trajectories, and a preference-based component that preserves inter-sample perceptual ordering through magnitude-aware pairwise and triplet comparisons. Trained with GRPO and a two-stage Exploration-to-Stability fine-tuning strategy, PreResQ-R1 enables MLLMs to perform human-aligned, chain-of-thought reasoning about visual fidelity, achieving both high correlation accuracy and robust cross-domain generalization.
\end{enumerate}
\end{document}